%% file: main.tex
\documentclass{article}
\usepackage{float}
\usepackage[margin=1.0in]{geometry}
\usepackage{abstract}

\input{macros}

\hypersetup{
  colorlinks=true,
  urlcolor=BrickRed,
  citecolor=ForestGreen,
  linkcolor=BrickRed
} 
\title{\raisebox{-0.05\height}{\includegraphics[height=0.8em]
{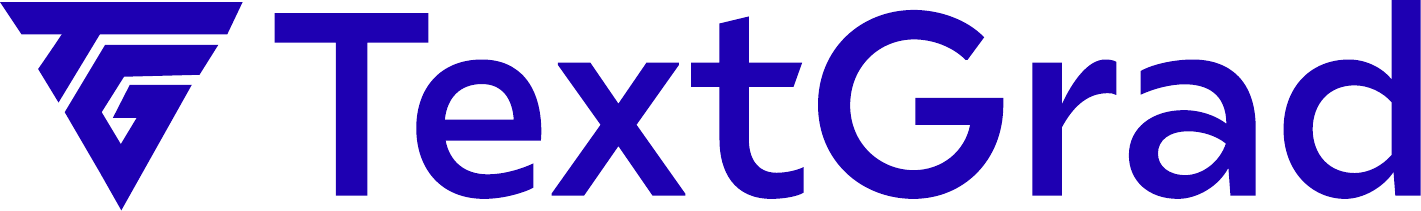}}\,: Automatic ``Differentiation'' via Text }
\author{\name Mert Yuksekgonul$^{1}$\thanks{Co-first authors.} \email merty@stanford.edu \\
\name Federico Bianchi$^{1}$\textsuperscript{*} \email fede@stanford.edu \\
\name Joseph Boen$^{2}$\textsuperscript{*} \email tboen@stanford.edu \\
\name Sheng Liu$^{2}$\textsuperscript{*} \email shengl@stanford.edu \\
\name Zhi Huang$^{2}$\textsuperscript{*} \email zhihuang@stanford.edu \\
\name Carlos Guestrin$^{1,3}$ \email guestrin@stanford.edu \\
\name James Zou$^{1,2,3}$ \email jamesz@stanford.edu \\
$^{1}$Department of Computer Science, Stanford University\\
$^{2}$Department of Biomedical Data Science, Stanford University\\
$^{3}$Chan Zuckerberg Biohub \\ 
\email Correspondence: merty@stanford.edu and jamesz@stanford.edu \\
\email \begin{normalsize}
\begin{center} \href{https://github.com/zou-group/textgrad}{\raisebox{-0.1\height}{\includegraphics[height=1em]
{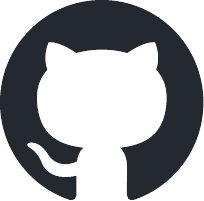}}
\textcolor{logocolor}{Repository and Tutorials}}\end{center}
\end{normalsize}
}

\begin{document}

\fancyhead[L]{\raisebox{-0.1\height}{\includegraphics[height=0.9em]
{figures/tg_logo_full.pdf}}}

\fancyhead[C]{\href{https://github.com/zou-group/textgrad}{\raisebox{-0.1\height}{\includegraphics[height=1em] {figures/github-mark.pdf}}}}
\fancyhead[R]{Automatic ``Differentiation'' via Text}
\setlength{\headheight}{13pt}
\pagestyle{fancy}

\newcounter{suppfigure}
\newcounter{supptable}
\makeatletter
\newcommand\suppfigurename{Supplementary Figure}
\newcommand\supptablename{Supplementary Table}
\newcommand\suppfigureautorefname{\suppfigurename}
\newcommand\supptableautorefname{\supptablename}
\let\oldappendix\appendix
\renewcommand\appendix{%
    \oldappendix
    \setcounter{figure}{0}%
    \setcounter{table}{0}%
    \renewcommand\figurename{\suppfigurename}%
    \renewcommand\tablename{\supptablename}%
}
\makeatother

\maketitle

\renewcommand{\abstractnamefont}{\normalfont\normalsize\bfseries}
\renewcommand{\abstracttextfont}{\normalfont\normalsize}

\renewcommand{\abstractname}{Abstract}
\begin{abstract}
\abstracttextfont
\input{sections/0_abstract}
\end{abstract}

\section{Introduction}
\input{sections/1_intro}

\begin{figure}[hbtp]
    \centering
\includegraphics[width=\textwidth]{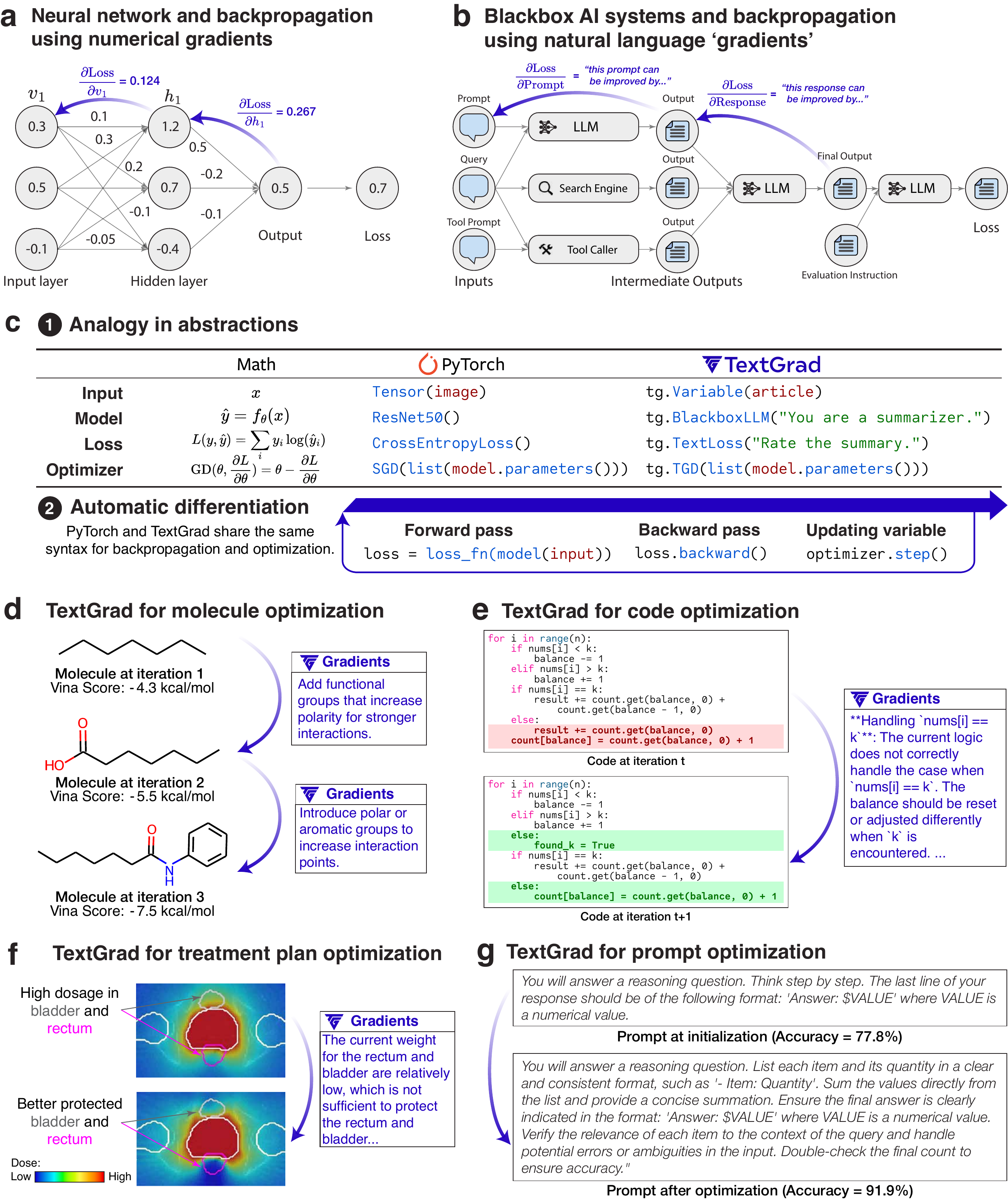}
    \caption{\textbf{Automatic ``Differentiation'' via Text~(a, b).} Backpropagation of gradients is the driving force of deep learning. We do not have gradients for compound systems of blackbox AI systems, but we can construct analogous backpropagation for text-based feedback, forming the basis of \textgrad. 
    \textbf{Abstractions in \textgrad~(c).} We share the same  abstractions and syntax as PyTorch to our framework generalizable and easy-to-learn. \textbf{Applications of \textgrad~(d,e,f,g).} In Section~\ref{sec:molecule}, we optimize molecular structures for properties such as druglikeness and protein binding affinity (d). In Section~\ref{sec:code-optimization}, we optimize solutions to coding problems (e). In Section~\ref{sec:treatment-plan}, we optimize radiotherapy treatment plans to improve patient outcomes (f). In Section~\ref{sec:prompt-optimization}, we optimize prompts to improve the reasoning of language models (g).}
    \label{fig:1}
\end{figure}

\section{\textgrad: Optimizing AI systems by backpropagating text feedback}
\label{sec:framework}
\input{sections/2_framework}

\section{Results}
\label{sec:benchmarks}

\input{sections/3_language_model_experiments}

\input{sections/4_scientific_applications}

\section{Related work}
\label{sec:rel_works}
\input{sections/5_related_works}

\section{Discussion}
\label{sec:discussion}
\input{sections/6_discussion}

\section*{Acknowledgements}
We would like to thank Duygu Yilmaz, Begum Ergun, Fatih Dinc, Yu Sun, Omar Khattab, Ian Covert, Kyle Swanson, Omer Faruk Akgun, Yusuf Efe, Kevin Y Wu, Eric Wu, Kailas Vodrahalli, Oscar Pastor Serrano, Patrick John Chia, Jacopo Tagliabue, Nitya Thakkar, Elana Simon, Pan Lu, Sabri Eyuboglu, Irena Gao, Lingjiao Chen, and members of the Zou Group for their support and comments on this work.

\clearpage

    \printbibliography
\clearpage
\appendix
\input{appendix/main}

\end{document}

%% file: macros.tex
\usepackage[dvipsnames,svgnames, table,xcdraw]{xcolor}
\usepackage{relsize}
\usepackage{tcolorbox}
\usepackage{listings}
\usepackage{subcaption}
\usepackage{algorithm}
\usepackage{algorithmic}
\usepackage{comment}
\usepackage{graphicx}
\usepackage{booktabs}
\usepackage{multirow}
\usepackage{todonotes}
\usepackage{enumitem}
\usepackage{url}
\usepackage{mathpazo}
\usepackage{amsmath,amssymb,amsthm,amsfonts,bm}
\usepackage[toc,page,header]{appendix}
\usepackage{fancyhdr}
\usepackage[backend=biber,style=nature,natbib=true,maxbibnames=99,minalphanames=3]{biblatex}
\usepackage[colorlinks=true]{hyperref}
\usepackage[T1]{fontenc}

\usepackage{auth_detailed}

\counterwithout{figure}{section}
\counterwithout{table}{section}
\definecolor{darkgoldenrod}{rgb}{0.72, 0.53, 0.04}
\definecolor{backgroundcolor}{RGB}{250, 250, 252}   
\definecolor{keywordcolor}{RGB}{30, 0, 178}       
\definecolor{stringcolor}{RGB}{204, 0, 102}        
\definecolor{numbercolor}{RGB}{0, 128, 128}        
\definecolor{emphcolor}{RGB}{30, 0, 178}            
\definecolor{commentcolor}{RGB}{0, 128, 0}       
\definecolor{basiccodecolor}{RGB}{61, 61, 61}       

\lstdefinestyle{customstyle}{
    backgroundcolor=\color{backgroundcolor},   
    commentstyle=\color{commentcolor},
    keywordstyle=\color{keywordcolor},
    numberstyle=\color{numbercolor},
    stringstyle=\color{stringcolor},
    basicstyle=\color{basiccodecolor}\ttfamily\footnotesize,
    breakatwhitespace=false,         
    breaklines=true,                 
    captionpos=b,                    
    keepspaces=true,                 
    numbers=left,     
    basicstyle=\color{basiccodecolor}\ttfamily\footnotesize,
    numbersep=5pt,             
    xleftmargin=2em,
    xrightmargin=2em,
    showspaces=false,                
    showstringspaces=false,
    showtabs=false,                  
    tabsize=1,
    frame=single,
    framesep=5pt,
    framexleftmargin=1.5em,
    framexrightmargin=1.5em,
    framextopmargin=1pt,
    framexbottommargin=1pt,
    aboveskip=10pt,
    belowskip=10pt,
    breaklines=true,
    breakautoindent=true,
    emph={textgrad, tg, Variable, MultipleChoiceTestTime,
    TextualGradientDescent, BlackboxLLM},             %
    emphstyle={\color{emphcolor}},
    extendedchars=true,
}

\input{editor}

\newcommand{\textllm}{\textcolor{RoyalPurple}{\text{LLM}}}
\newcommand{\textbackward}[1]{\textcolor{RoyalPurple}{\mathlarger{\nabla}_{\text{#1}}}}
\newcommand{\textbackwardchain}[1]{\textcolor{RoyalPurple}{\mathlarger{\nabla}_{\text{#1}}}}

\newcommand{\textparameter}[1]{\textcolor{ForestGreen}{\textbf{\text{#1}}}}

\newcommand{\textttgrad}[1]{\textcolor{ForestGreen}{\texttt{#1}}}

\definecolor{logocolor}{RGB}{30, 0, 178}                %
\newcommand{\textgrad}{{\textcolor{logocolor}{\textbf{\textsc{TextGrad}}}}}

\newcommand{\gpto}{\texttt{gpt-4o}}
\newcommand{\gptthreefive}{\texttt{gpt-3.5-turbo-0125}}

\newcommand{\textgradurl}{\url{https://github.com/zou-group/textgrad}}

\newcommand{\COMMENTT}[1]{\textcolor{ForestGreen}{\# \texttt{#1}}}

\definecolor{darkerlogocolor}{RGB}{20, 0, 145}  

\newtcolorbox{ttcolorbox}[1][]{colframe=darkerlogocolor, colback=darkerlogocolor!4!white, title=#1}

\newtcolorbox{apxtcolorbox}[1][]{colframe=black, colback=black!3!white, title=#1}

%% file: editor.tex
\usepackage[normalem]{ulem}
\usepackage{xcolor}

\newcommand{\scrcmd}{\textsuperscript}
\newcommand{\scr}[1]{\scrcmd{#1}}

%% file: sections/0_abstract.tex
AI is undergoing a paradigm shift, with breakthroughs achieved by systems orchestrating multiple large language models (LLMs) and other complex components. 
As a result, developing principled and automated optimization methods for compound AI systems is one of the most important new challenges.
Neural networks faced a similar challenge in its early days until backpropagation and automatic differentiation transformed the field by making optimization turn-key. 
Inspired by this, we introduce \textgrad, a powerful framework performing automatic ``differentiation'' via text. \textgrad~backpropagates textual feedback provided by LLMs to improve individual components of a compound AI system. 
In our framework, LLMs provide rich, general, natural language suggestions to optimize variables in computation graphs, ranging from  code snippets to molecular structures. 
\textgrad~follows PyTorch's syntax and abstraction and is flexible and easy-to-use. It works out-of-the-box for a variety of tasks, where the users only provide the objective function without tuning components or prompts of the framework.
We showcase \textgrad's effectiveness and generality across a diverse range of applications, from question answering and molecule optimization to radiotherapy treatment planning. Without modifying the framework, \textgrad~improves the zero-shot accuracy of GPT-4o in Google-Proof Question Answering from $51\%$ to $55\%$, yields $20\%$ relative performance gain in optimizing LeetCode-Hard coding problem solutions, improves prompts for reasoning, designs new druglike 
small molecules with desirable \textit{in silico} binding, and designs radiation oncology treatment plans with high specificity.
\textgrad~lays a foundation to accelerate the development of the next-generation of AI systems.

%% file: sections/1_intro.tex
There is an emerging paradigm shift in how AI systems are built, owing to the breakthroughs of Large Language Models~(LLMs)~\citep{brown2020language, achiam2023gpt, reid2024gemini, llama3modelcard, anthropic2024claude, bommasani2021opportunities}. The new generation of AI applications are increasingly compound systems involving multiple sophisticated components, where each component could be an LLM-based agent, a tool such as a simulator, or web search.
For instance, a system of LLMs communicating with symbolic solvers can solve olympiad-level math problems~\citep{trinh2024solving}; a system of LLMs using search engines and code interpreter tools performs comparably to  human competitive programmers~\citep{li2022competition} and are solving real-world GitHub issues~\citep{yang2024sweagent}. However, many of these breakthroughs came from systems that are hand-crafted by experts in the domain of application and are tweaked through heuristics. Therefore, developing principled and automated ways to optimize AI systems is one of the most crucial challenges for building compound systems with LLMs, and is necessary for unlocking the power of AI~\citep{khattab2024dspy, zaharia2024shift, zhou2022large}.

For the past 15 years, many advances in AI have relied on artificial neural networks and differentiable optimization~\citep{krizhevsky2012imagenet, jumper2021highly, fawzi2022discovering, mankowitz2023faster, merchant2023scaling, trinh2024solving}. Different parts of neural networks~(e.g., two artificial neurons) communicate through differentiable functions like matrix multiplications~\citep{goodfellow2016deep}. Therefore, using numerical gradients and backpropagation~\citep{rumelhart1986learning}, which provide the direction to adjust each parameter to improve a model, has been the natural way to train AI models. Flexible automatic differentiation frameworks implementing backpropagation~\citep{jia2014caffe, bergstra2010theano, abadi2016tensorflow, paszke2019pytorch, collobert2002torch} have been indispensible to the development of AI \emph{models}.

To optimize the new generation of AI \emph{systems}, we introduce \textgrad, automatic differentiation via text. Here we use differentiation and gradients as a metaphor for textual feedback from LLMs. In this framework, each AI system is transformed into a computation graph, where variables are inputs and outputs of complex~(not necessarily differentiable) function calls. The feedback to the variables~(dubbed `textual gradients'~\citep{pryzant-etal-2023-automatic}) are provided in the form of informative and interpretable natural language criticism to the variables; describing how a variable should be changed to improve the system. The gradients are propagated through arbitrary functions, such as LLM API calls, simulators, or external numerical solvers.

We demonstrate the power of our framework in a diverse set of domains, ranging from question answering benchmarks to radiotherapy treatment plan optimization and molecule generation~(Fig.~\ref{fig:1}). LLMs can provide very rich, legible, and expressive natural language gradients to variables in this wide range of domains, such as proposing modifications to molecules, prompts to other LLMs, and code snippets. Our framework is built on the assumption that the current state-of-the-art LLMs are able to reason about individual components and subtasks of the system that it tries to optimize. We demonstrate the flexibility of \textgrad~with the following results:
\begin{enumerate}[leftmargin=1cm, itemsep=-0.5ex, topsep=0pt]
    \item \textbf{Coding:} In Section~\ref{sec:code-optimization}, we optimize solutions to difficult coding problems from LeetCode~\citep{reflexion}, where we boost the performance of \gpto~and best existing method by $20\%$ relevant performance gain.
    \item \textbf{Problem Solving:} In Section~\ref{sec:solution-optimization}, we optimize solutions to complex scientific questions to improve the zero-shot performance of GPT-4o. For instance, in Google-Proof Question Answering~\citep{rein2023gpqa} benchmark, we improve the zero-shot accuracy from $51\%$ to $55\%$ by refining the solutions at test-time.
    \item \textbf{Reasoning:} In Section~\ref{sec:prompt-optimization}, we optimize prompts to improve the LLM performance, where we push the performance of GPT-3.5 close to GPT-4 in several reasoning tasks.
    \item \textbf{Chemistry:} In Section~\ref{sec:molecule}, we design new small molecules with desirable druglikeness and \textit{in silico} binding affinity to drug targets. 
    \item \textbf{Medicine:} In Section~\ref{sec:treatment-plan}, we optimize radiation treatment plans for prostate cancer patients to achieve desirable target dosage and reduce side effects. 
\end{enumerate}

Our results in a broad set of applications demonstrate the promise of \textgrad ~to automatically optimize compound AI systems via backpropagation of text feedback. 
To accelerate progress in this direction, we open-source our framework at \textgradurl.

%% file: sections/2_framework.tex
Below we first describe a example to demonstrate what \textgrad~looks like for a system made of two LLM calls, and then give the more general form for arbitrarily complex systems. 

\subsection*{Warmup: system with two LLM calls}

\paragraph{Example computation graph.} In traditional automatic differentiation, we compute gradients that provides a direction that would improve the variable with respect to a downstream loss with the chain-rule. For a simple analog, let us look at the simple system:
\begin{align}
    \text{Prediction} & = \textllm(\textparameter{Prompt} + \text{Question}), \\
    \text{Evaluation} & = \textllm(\text{Evaluation Instruction} + \text{Prediction}),
\end{align}
where we denote the free parameter to optimize, the prompt,  with $\textparameter{green}$, and we use $+$ to denote concatenation of two strings, and use $\textllm(x)$ to give $x$ as a prompt to a language model to collect the response. We will use the chain notation: \begin{equation}
\textparameter{Prompt} + \text{Question} \xrightarrow{\textllm} \text{Evaluation Instruction} + \text{Prediction} \xrightarrow{\textllm} \text{Evaluation}, \label{eq:simple}\end{equation} as another way to denote this system, where we make one $\textllm$ call to generate a $\text{Prediction}$ for a $\text{Question}$ using a $\textparameter{Prompt}$, and another $\textllm$ call to evaluate this $\text{Prediction}$.

\noindent \paragraph{Example gradient computation.} In this example system, to improve the $\textparameter{Prompt}$ with respect to the evaluation, we instantiate an analog of the backpropagation~\citep{rumelhart1986learning} algorithm through the following:

\begin{figure}[H]
\begin{ttcolorbox}[Gradient computation for the simple graph: $
\textbf{\textparameter{Prompt}}\xrightarrow{\textcolor{white}{\text{LLM}}} \text{Prediction} \xrightarrow{\textcolor{white}{\text{LLM}}} \text{Evaluation}$]
\begin{align}
    \frac{\partial \text{Evaluation}}{\partial \text{Prediction}} & = \textbackward{\text{LLM}}(\text{Prediction}, \text{Evaluation}), \\ 
    \frac{\partial \text{Evaluation}}{\partial \textparameter{Prompt}} = \frac{\partial \text{Evaluation}}{\partial \text{Prediction}} \circ \frac{\partial \text{Prediction}}{\partial \textparameter{Prompt}} & = \textbackwardchain{\text{LLM}}(\textparameter{Prompt},  \text{Prediction}, \frac{\partial \text{Evaluation}}{\partial \text{Prediction}}),
\end{align}

\noindent where we use $\textbackwardchain{\text{LLM}}$ for the gradient operator when the forward function is an LLM call.\footnote{Note that we overload the notation to denote both for cases when the output variable does not have successors~(e.g., $\textbackwardchain{\textllm}(\text{Prediction}, \text{Evaluation})$) and when it has successors, and thus gradients~(e.g., $\textbackwardchain{\text{LLM}}(\textparameter{Prompt},  \text{Prediction}, \frac{\partial \text{Loss}}{\partial \text{Prediction}})$).} In particular, this function returns natural language feedback such as `\textit{This prediction can be improved by\ldots}' where the feedback describes how to modify the variable to improve the downstream objective, analogous to gradients in optimization. We first collect feedback to the $\text{Prediction}$ variable using the evaluation. Then, given this feedback and the ($\textparameter{Prompt} \xrightarrow{\textllm} \text{Prediction}$) call, we collect the feedback on the $\textparameter{Prompt}$.

\end{ttcolorbox}
\end{figure}

\noindent  Below is a flexible way to instantiate $\textbackwardchain{\text{LLM}}$ to collect feedback for a simple system like $
x \xrightarrow{\textllm} y \xrightarrow{\textllm} \mathcal{L} $:

\begin{figure}[H]
\begin{ttcolorbox}[Example implementation for the gradient operator]

\begin{align}
\frac{\partial \mathcal{L}}{\partial x} = \textbackwardchain{\text{LLM}}(x, y, \frac{\partial \mathcal{L}}{\partial y}) \triangleq & \ "\texttt{Here is a conversation with an LLM: \{x|y\}.}" \\ 
& \ + \nonumber \\ 
& \ \textllm( \texttt{Here is a conversation with an LLM: \{x|y\}.} \nonumber \\ 
& \ \phantom{\textllm(} \texttt{Below are the criticisms on \{y\}:} \nonumber \\ 
& \ \phantom{\textllm(} \left \{\frac{\partial \mathcal{L}}{\partial y} \right \} \nonumber \\ 
& \ \phantom{\textllm(} \texttt{Explain how to improve \{x\}.}), \nonumber
\end{align}
where the gradient object is a combination of the context in which the variable appears and the feedback obtained from an LLM\footnote{The exact prompts we use are different to ensure generality and flexibility; we use these examples only for exposition.}, defined analogously to \citet{pryzant-etal-2023-automatic}. Note that this operator does not depend on, e.g., the application domain. Once implemented, the gradient operator for LLM calls are fixed throughout the framework for all applications.

\end{ttcolorbox}
\end{figure}

\noindent \paragraph{Example optimizer.} In standard gradient descent, the current value of the variable is combined with the gradients through subtraction, e.g.:

\begin{equation}
\theta_{\text{new}} = \textcolor{RoyalPurple}{\text{GradientDescent.step}}(\theta, \frac{\partial \mathcal{L}}{\partial \theta}) = \theta - \frac{\partial \mathcal{L}}{\partial \theta}.
\end{equation}

Continuing the gradient-based optimization analogy, we use \textcolor{RoyalPurple}{\emph{Textual} Gradient Descent~(\text{TGD})}:

\begin{figure}[H]
\begin{ttcolorbox}[Updating the $\textbf{Prompt}$ variable in the simple graph via TGD.]
\begin{equation}
\textparameter{Prompt}_{\textparameter{new}} = \textcolor{RoyalPurple}{\text{TGD.step}} (\textparameter{Prompt}, \frac{\partial \text{Evaluation}}{\partial \textparameter{Prompt}}).
\end{equation}

to update the parameters, in this case, the $\textparameter{Prompt}$. In particular, given the current variable and the gradients~(feedback) we collected for this variable, the optimizer seeks to update this variable.
\end{ttcolorbox}
\end{figure}

A concrete way to instantiate \textcolor{RoyalPurple}{\text{TGD}} is the following:

\begin{figure}[H]
\begin{ttcolorbox}[Example implementation of one TGD iteration]
\noindent 
\begin{align}
    x_{\textrm{new}} = \textcolor{RoyalPurple}{\text{TGD.step}}(x, \frac{\partial \mathcal{L}}{\partial x}) \triangleq & \ \textllm(\texttt{Below are the criticisms on \{x\}:} \\ & \ \phantom{\textllm(} \left \{ \frac{\partial \mathcal{L}}{\partial x} \right \} \nonumber \\ & \ \phantom{\textllm(} \texttt{Incorporate the criticisms, and produce a new variable.}). \nonumber
\end{align}
where $x$ is the variable we would like to improve, and $\frac{\partial \mathcal{L}}{\partial x}$ is the feedback we obtained for the variable during the backward pass\footnote{The exact prompts we use are different to ensure generality and flexibility; we use these examples only for exposition.}.  Similar to the gradient operator, this function also does not depend on the domain of application, and TGD implementation is the same across all uses of the framework.
\end{ttcolorbox}
\end{figure}

\subsection*{The general case}

\noindent The abstraction readily applies to arbitrarily complex systems. Define a computation graph by
\begin{equation}
    v = \textcolor{RoyalPurple}{f_v}(\text{PredecessorsOf}(v)) \ \ \forall v \in \mathcal{V}
\end{equation}
where $v$ is a variable in the graph, $\mathcal{V}$ is the set of all variables in the graph, and $\text{SuccessorsOf}$ returns the successors and $\text{PredecessorsOf}$ returns the predecessors of a variable. Generally speaking, the value of $v$ can be unstructured data, such as natural language text or images. For most of the results and exposition in this paper, $v$ is natural language text.

Further, let us have $\textcolor{RoyalPurple}{f_v}$ as the transformation that consumes a set of variables and produces the variable $v$. For instance, we can use an $\textllm$ or $\textcolor{RoyalPurple}{\text{Numerical Simulator}}$ as a transformation. Since different functions will have different ways to compute gradients and collect feedback for, we will generally use $\textbackward{f}$ to denote the gradient function for a function $\textcolor{RoyalPurple}{f}$. For the sake of exposition, we will omit the subscript when the function is obvious.

The gradients are computed by 

\begin{equation}
\frac{\partial \mathcal{L}}{\partial v} = \underset{w \in \text{SuccessorsOf}(v)}{\bigcup} \mathlarger{\textbackwardchain{f}}\left(v, w,  \frac{\partial \mathcal{L}}{\partial w}\right),
\label{eq:recursive}
\end{equation}

\noindent where we collect the set of gradients from all successors of $v$. Intuitively, we get feedback from every context in which a variable $v$ was used, and aggregate them.

Equation~\ref{eq:recursive} recursively computes the gradients of the downstream objective with respect to the desired variables $v$ in the graph. The $\textbackward{f}$ function takes as input the gradients of $\mathcal{L}$ with respect to the successors of a given variable $v$, the value of the variable $v$ and the successors themselves. Note that the final gradient variable comprises a set of contexts and criticisms for any place a variable was used in.

Finally, to update any desired variable $v$ in the graph, we can use an optimizer:
\begin{equation}
v_{\textrm{new}} = \textcolor{RoyalPurple}{\text{TGD.step}}\left(v, \frac{\partial \mathcal{L}}{\partial v}\right),
\end{equation}
which updates the value of $v$ based on its current value and the gradients. For a computation graph where there are $n$ edges, each iteration of optimization performs at most $n$ additional language model calls to compute gradients~(1 call using the gradient operator for each edge in the computation graph). For implementations of operations in \textgrad, see Appendix~\ref{appendix:textgrad-operations}. 

\subsection*{Objective functions}

In numerical optimization and automatic differentiation, the objective function is typically a differentiable function, such as mean squared error or cross entropy. In \textgrad, the objective can be a complex and potentially nondifferentiable function, where the domain and codomain of the function can be unstructured data. This choice adds important generality and flexibility to the framework. For instance,  we demonstrate that the objective can be specified in natural language text and computed by prompting a language model~(\S\ref{sec:prompt-optimization}), an output of a code interpreter running unit tests~(\S\ref{sec:code-optimization}), or outputs of molecular simulation engines~(\S\ref{sec:molecule}). For instance, a simple loss function for a code snippet can be the following:

\begin{align}
    \text{Loss}(\textparameter{code},  \text{target goal}) = & \textllm(\text{Here is a code snippet:}\{\textparameter{code}\}. \label{eq:examples-code}\\ & \phantom{\textllm(}\text{Here is the goal for this snippet:}\{\text{target goal}\}. \nonumber \\ & \phantom{\textllm(}\text{Evaluate the snippet for correctness and runtime complexity.}), \nonumber
\end{align}

\noindent where we can use this evaluation signal to optimize the code snippet, powered by the well-documented ability of LLMs to simulate human feedback, self-evaluate, and self-improve~\citep{li2023alpacaeval, bai2022training, madaan2024self, stiennon2020learning, yuan2024self, dubois2024alpacafarm, khattab2024dspy}.

\subsection*{Instance vs Prompt Optimization}
There are two classes of optimization problems we explore. 

\textbf{In instance optimization,} we directly treat a solution to a problem---e.g., a code snippet, the solution to a problem or a molecule---as an optimization variable. For instance, in Equation~\ref{eq:examples-code}, we have a code instance that we would like to improve at test time. Our framework produces the gradients for and directly optimizes the \textparameter{code} variable.

\textbf{In prompt optimization,} the goal is to find a prompt that improves the performance of an LLM across multiple queries for a task. For example,  we may want to find a system prompt to an LLM that improve the performance on mathematical reasoning questions~(see Section~\ref{sec:prompt-optimization} for examples). In particular, we want the system prompt to generalize, in contrast to instance optimization where the only goal is to improve the solution for a given query at test time. 

Crucially, both types of problems can be solved without hand-crafting the framework.

\subsection*{Optimization Techniques}
Automatic differentiation is a strong analogy for \textgrad~and provides conceptual support for optimization techniques implemented in the framework. Below, we describe some examples.

\textbf{Batch Optimization:} We implement stochastic minibatch gradient descent~\citep{bottou2010large, goodfellow2016deep} for prompt optimization. Specifically, after doing a forward pass with multiple instances in a batch and evaluate individual loss terms, we use the $\texttt{tg.sum}$ function to sum the losses~(akin to $\texttt{torch.sum}$). In the backward pass, gradients on variables through individual loss terms are  concatenated, mirroring backpropagating through addition.

\textbf{Constrained Optimization:} We use natural language constraints, building on constrained optimization as an analogy~\citep{boyd2004convex}. In particular, we use natural language constraints~(e.g., such as \textit{`The last line of your response should be of the following format: 'Answer: \$LETTER' where LETTER is one of ABCD.''}) to guide the behavior of the optimizer. We observe that thanks to instruction-tuning~\citep{ouyang2022training, wei2022finetuned}, language models can follow these simple constraints, although their reliability can reduce with too many constraints~\citep{yuksekgonul2024attention, abdin2024kitab}.

\textbf{Momentum:} We use the analogy to momentum in gradient descent~\citep{polyak1964some, sutskever2013importance}. When optimizing a variable, the $\textcolor{RoyalPurple}{\text{TGD}}$ optimizer can optionally see the earlier iterations of the variable when making the update.  

\noindent For more details on optimization techniques that are implemented in \textgrad, see Appendix~\ref{appendix:optimizer-extensions}.

%% file: sections/3_language_model_experiments.tex
We demonstrate the flexibility of \textgrad~in a diverse array of applications.  In \S~\ref{sec:code-optimization}, we optimize code snippets to solve hard coding problems from LeetCode. In  \S~\ref{sec:solution-optimization}, we optimize solutions to science questions. In \S~\ref{sec:prompt-optimization} we optimize prompts to improve the reasoning of LLMs. In \S~\ref{sec:molecule}, we optimize chemical structures for improved molecular properties. In \S~\ref{sec:treatment-plan}, we optimize treatment plans for prostate cancer patients.

\subsection{Code optimization}
\label{sec:code-optimization}
Code optimization is a hallmark use case of instance optimization. Here, the goal is to optimize some code to improve e.g., correctness or runtime complexity. We often have a computation graph like the following:

\begin{equation}
    \text{Code-Refinement Objective} = \textcolor{RoyalPurple}{\textllm}(\text{Problem} + \textparameter{Code} + \text{Test-time Instruction + Local Test Results})
\label{eq:self-supervision-code-opt}
\end{equation}

\noindent where we optimize the $\textparameter{Code}$ to solve a given \text{Problem} with limited, local test supervision and self-evaluation through a test-instruction asking to critique the current iteration of the code. Figure~\ref{fig:1}e shows an example for the problem \textit{You are given an array nums of size n consisting of distinct integers from 1 to n and a positive integer k. Return the number of non-empty subarrays in nums that have a median equal to k.} The first solution proposed by \gpto~does not pass the tests. \textgrad~identifies an edge case in the first solution and provides a suggestion on how to improve it. The optimized implementation passes all tests.

\textbf{Task:}
We use the LeetCode Hard dataset~\citep{reflexion} to benchmark code optimization. LeetCode is an online platform that offers coding practice questions in preparation for technical interviews. The LeetCode Hard dataset contains examples of \emph{hard} coding problems that are meant to be challenging for both humans and language models, where the success metric is Completion Rate, i.e., passing all test cases for a given problem~(GPT-4 reportedly achieved a 7\% completion rate~\cite{reflexion}). LeetCode test cases are not public, and thus, after generation, the code has to be submitted to the LeetCode platform for evaluation on the unseen test cases. This makes the platform more suitable to evaluate the performance of language models.

\textbf{Baseline:} 
Reflexion~\citep{reflexion} is the state-of-the-art method on the LeetCode Hard dataset. Their approach prompts an LLM to self-reflect on code snippets and the errors that were generated at test time using candidate unit tests. Given the self-reflection, the LLM is prompted again to provide an updated piece of code, conditioned on the self-reflection and the errors. We ran Reflexion on LeetCodeHard using \gpto~using 1 in-context demonstration to guide the behavior~(one-shot). In addition to Reflexion, we also run a zero-shot baseline using \gpto~mimicking the same zero-shot baseline described in~\cite{reflexion}. In comparison, \textgrad~runs in a zero-shot setting, without any demonstrations.

\textbf{Results:}
Existing results~\citep{reflexion} showed a 7\% pass rate for GPT-4 zero-shot and 15\% for GPT-4 with Reflexion. We show that these results have now been boosted to 23\% for \gpto~zero-shot and 31\% when Reflexion is used. With \textgrad, we can optimize solutions to achieve a performance of 36\%. These improvements are more impressive considering that Reflexion was ran with in-context demonstrations and \textgrad~did not use any demonstrations (i.e. zero-shot).

\begin{ttcolorbox}[Code Refinement Objective]
$\textllm($"You are an intelligent assistant used as an evaluator, and part of an optimization system. 
You will analyze a code implementation for a coding problem and unit test results. 
The code will be tested with harder tests, so do not just check if the code passes the provided tests.
Think about the correctness of the code and its performance in harder test cases.
Give very concise feedback.
Investigate the code problem and the provided implementation. 
For each failed unit test case, start analyzing it by saying "The code did not pass this test because...". 
Explain why the current implementation is not getting the expected output. Do not provide a revised implementation. Carefully 
suggest why there are issues with the code and provide feedback. \\
\{Test-time Instruction\} \\ 
**The coding problem:**

\{Problem\}

**Code generated that must be evaluated for correctness and runtime performance**

\{\textparameter{Code}\}

**The test results:**

\{Local test Results\}
\end{ttcolorbox}

\begin{table}[!h]
\caption{Code optimization for LeetCode Hard using \gpto. Results are averaged over 5 seeds.}
    \begin{tabular}{@{}lll@{}}
    \toprule
    \textbf{Task} & \textbf{Method}   & \textbf{Completion Rate}   \\ \midrule
    \multirow{3}{*}{LeetCode Hard~\citep{reflexion}} & Zero-shot~\cite{reflexion} & $0.26$ \\ & Reflexion (1 demonstration, 5 iterations)~\citep{reflexion} & $0.31 \pm 0.012$ \\
    & \textgrad~ (0 demonstrations, 5 iterations) &  $\mathbf{0.36} \pm 0.018 $ \\
    \bottomrule
    \end{tabular}
    \centering
    \label{tab:code-optimization}
\end{table}

\subsection{Solution optimization by test-time training to improve problem solving}
\label{sec:solution-optimization}

Here, we focus on the task of \emph{solution optimization} via \textgrad. 
In solution optimization, the goal is to improve the solution to a complex problem, such as a question about quantum mechanics or organic chemistry. In particular, we often have a computation graph like the following:

\begin{equation}
    \text{Solution Refinement Objective} = \textcolor{RoyalPurple}{\textllm}(\text{Question} + \textparameter{Solution} + \text{Test-time Instruction})
\label{eq:self-supervision-solution-opt}
\end{equation}

where the parameter we optimize is the $\textparameter{Solution}$, and the loss function is obtained by an evaluation of the solution, e.g., with an LLM. At each iteration, the $\textllm$ is prompted with the question, current solution, and some test-time instruction asking to critique or investigate the current iteration. Over the optimization trajectory, the solution is refined using this test-time self-evaluation. More generally, this idea is known as test-time training~\citep{sun2020test, sun2023learning}, where a machine learning model is trained on a test instance at test-time, often with a self-supervised objective.  Similarly, recent work have shown the merits of self-refinement also for reasoning tasks~\citep{madaan2024self, reflexion, yao2023react}. In particular, even though an LLM may not get the answer to a question or the solution to a problem right at first attempt, it can improve the response through iterative refinement.

For instance, the objective function for the refinement looks like the following:

\begin{ttcolorbox}[Solution Refinement Objective]
$\textcolor{RoyalPurple}\textllm($"Below is a multi-choice question and a prediction. You are a critical and creative scientist. Your job is to investigate the prediction. Critically go through reasoning steps, and see if there is a reason why the prediction could be incorrect.

Use the Janusian Process, think about whether alternative answers could be true.

Question: $\{\text{Question}\}$

Answer by the language model: $\{\textparameter{Solution}\}$")\end{ttcolorbox}

\noindent Below is a representative implementation of solution optimization in \textgrad.

\noindent\begin{minipage}{\textwidth}
\centering
\begin{lstlisting}[language=Python, caption=A representative implementation of solution optimization in \textgrad.]
# Assume we have the test_dataset 
question, answer = test_dataset[i]
# Initialize the test time loss function
# This has the system prompt provided above as `Solution Refinement Objective'
test_time_loss_fn = MultipleChoiceTestTime()
# Get a zero-shot solution from an LLM
solution: Variable = zero_shot_llm(question)
# Optimize the solution itself
optimizer = tg.TextualGradientDescent(parameters=[solution])

for iteration in range(max_iterations):
    optimizer.zero_grad()
    # Note how the loss is self-supervised (does not depend on any ground truth.)
    loss = test_time_loss_fn(question, solution)
    # Populate the gradients, and update the solution
    loss.backward()
    optimizer.step()

\end{lstlisting}
\end{minipage}

\textbf{Task:} Google-proof Question Answering~(GPQA)~\cite{rein2023gpqa} is a recent benchmark where challenging multiple-choice questions in physics, biology, and chemistry are created and labeled by domain experts who have or are pursuing PhD degrees. In this benchmark, experts and skilled non-experts are reported to achieve $81\%$ and $22\%$ accuracy respectively, demonstrating the difficulty of the questions. Importantly, this is a benchmark where performance has not yet saturated, where to our knowledge, the best reported results, achieved by \gpto, gets $53.6\%$ accuracy in the diamond subset. We also use two challenging subsets~(Machine Learning and College Physics) of the MMLU~\citep{hendrycks2021measuring} question answering benchmark that is used to track the progress of language modeling and whether LLMs reached human-level performance.  Here the expert human accuracy on average is around $90\%$. For the details of the question format and prompts, please see Appendix~\ref{appendix:solution-optimization}.

\textbf{Method:}  We report two baselines. First, the reported results in the \gpto~release document states $53.6\%$ accuracy. However, their official implementation uses a temperature of $0.5$ for generations, thus we also test \gpto~with temperature $0$ and Chain-of-Thought~(CoT)~\citep{kojima2022large, wei2022chain} prompting provided in the official implementation\footnote{We do this to minimize randomness, however, \href{https://152334h.github.io/blog/non-determinism-in-gpt-4/}{ discussions} claim there may be other sources of non-determinism with \texttt{gpt-4}.}. For \textgrad, we perform $3$ iterations of test-time updates~(i.e. update the solution three times) and perform majority voting across all solutions to get the final answer. We use string-based metrics to compute the final accuracy of each answer.

\textbf{Results:} With \textgrad, we improve the performance of \gpto~in the challenging question-answering tasks and report the results in Table~\ref{tab:solution-optimization}. To our best knowledge, $55\%$ is the best known result in the GPQA dataset so far. Similarly, we improve the performance in MMLU subsets from $85.7\%$ to $88.4\%$~(Machine Learning) and $91.2\%$ to $95.1\%$~(College Physics). These results show that by spending more computation at test-time through \textgrad~self-refinement, we can improve the question answering performance of even the most capable models.

\begin{table}[!h]
\caption{Solution optimization for zero-shot question answering with \gpto.}
    \begin{tabular}{@{}lll@{}}
    \toprule
    \textbf{Dataset} & \textbf{Method}   & \textbf{Accuracy~($\%$)} \\ \midrule
    \multirow{3}{*}{Google-proof QA~\citep{rein2023gpqa}} & CoT~\citep{wei2022chain, kojima2022large} & $51.0$   \\ 
    & Best reported~\citep{openai2024hello} & $53.6$   \\ 
    & \textgrad & $\mathbf{55.0}$ \\ \midrule
    \multirow{2}{*}{MMLU-Machine Learning~\citep{hendrycks2021measuring}} & CoT~\citep{wei2022chain, kojima2022large} & $85.7$   \\ 
    & \textgrad & $\mathbf{88.4}$ \\ \midrule
    \multirow{2}{*}{MMLU-College Physics~\citep{hendrycks2021measuring}} & CoT~\citep{wei2022chain, kojima2022large} & $91.2$   \\ 
    & \textgrad & $\mathbf{95.1}$ \\ 
    \bottomrule
    \end{tabular}
    \centering
    \label{tab:solution-optimization}
\end{table}

\subsection{Prompt optimization for reasoning}
\label{sec:prompt-optimization}

While LLMs demonstrate an impressive performance in reasoning tasks, their performance can be sensitive to the prompt used to guide their behavior. In particular, with the right choice of a prompt, their performance can be significantly improved~\citep{liu2023pre}. In prompt optimization, the goal is to find a prompt or an instruction to guide the behavior of an LLM, such that it performs well on a given task. In particular, we often have a computation graph like the following:

\begin{align}
    \text{Answer} & = \textllm(\textparameter{Prompt}, \text{Question}) \nonumber \\
    \text{Evaluation Metric} & = \text{Evaluator}(\text{Answer}, \text{Ground Truth})
\label{eq:prompt-fw}
\end{align}

\noindent where we have a $\text{Question}$ for the task, an $\text{Answer}$ to the question, and an $\text{Evaluation Metric}$ indicating the quality of the output given the ground truth answer. For instance, for a question-answering task, the evaluation metric would be the accuracy of the answer, evaluated e.g. using string-based metrics.

Here, given a handful of training examples to optimize a $\textparameter{Prompt}$, the goal is to maximize the performance of an LLM on the given task. In our experiments, our goal is to improve the performance of a weaker and cheaper model~(e.g., \gptthreefive) using the feedback generated by stronger models (e.g., \gpto). This is useful in practice because by paying a fixed cost to optimize a $\textparameter{Prompt}$, the prompt-optimized weaker model can be used with cheaper inference costs instead of using the strong and more expensive model.

\noindent\begin{minipage}[!h]{\textwidth}
\begin{lstlisting}[language=Python, caption=A simple implementation of prompt optimization with \textgrad.]
# Initialize the system prompt
system_prompt = tg.Variable("You are a helpful language model. Think step by step.", 
                            requires_grad=True,
                            role_description="system prompt to the language model")

# Set up the model object 'parameterized by' the prompt.
model = tg.BlackboxLLM(system_prompt=system_prompt)

# Optimize the system prompt
optimizer = tg.TextualGradientDescent(parameters=[system_prompt])

for iteration in range(max_iterations):
    batch_x, batch_y = next(train_loader)
    optimizer.zero_grad()
    # Do the forward pass
    responses = model(batch_x)
    losses = [loss_fn(response, y) for (response, y) in zip(responses, batch_y)]
    total_loss = tg.sum(losses)
    # Perform the backward pass and compute gradients
    total_loss.backward()
    # Update the system prompt
    optimizer.step()
\end{lstlisting}
\label{fig:prompt_optimization_code}
\end{minipage}

\noindent Here, we use \textgrad~in a minibatch stochastic gradient descent setting~\citep{goodfellow2016deep, bottou2010large}. In particular, at each iteration, we use a few training examples to run the forward pass in Equation~\ref{eq:prompt-fw}. A pseudocode and short implementation can be found in the snippet above. Full details of prompt optimization can be found in Appendix~\ref{appendix:prompt-optimization}. Unlike instance optimization, where \textgrad~ tries to optimize each individual solution, the goal here is to optimize a single prompt that works well across all the questions in a benchmark. 

\textbf{Tasks:} We explore prompt optimization in multiple datasets, including the two standard reasoning tasks~(Object Counting and Word Sorting) from Big Bench Hard~(randomly split into 50/100/100 train/validation/test samples)~\citep{suzgun-etal-2023-challenging, srivastava2023beyond}, GSM8k grade-school math problem solving~\citep{cobbe2021gsm8k}~(using train/validation/test splits from DSPy). For GSM8k and Object Counting, we use a string-based exact match metric to quantify accuracy (i.e. whether the final number provided in the response is the same as the ground truth answer). For Word Sorting, we use an LLM to compare the response and the ground truth answer. We give more details about the tasks, prompts, and example queries in Appendix~\ref{appendix:prompt-optimization-tasks}.

\textbf{Methods:} We explore improving the performance of \gptthreefive~using \gpto~\citep{openai2024hello} to provide feedback during backpropagation. In particular, while the forward model that performs the reasoning is \gptthreefive, we use \gpto~to provide the feedback and improve the prompt. We use a batch size of $3$ with $12$ iterations, i.e., the model sees $36$ training examples in total, sampled randomly with replacement. After each iteration, we run a validation loop with the validation set of the datasets, and if the performance is better than the previous iteration we update the $\textparameter{Prompt}$.

\textbf{Baselines:} We have two main baselines:
\begin{enumerate}[leftmargin=0.5cm, itemsep=-0.5ex, topsep=0pt]
    \item \textbf{Zero-shot Chain-of-Thought~(CoT)~\citep{wei2022chain, kojima2022large}:} We initialize all prompts as zero-shot CoT prompt, where a model is instructed to `Think step-by-step' to explain its reasoning before giving its answer. This strategy is well-known to be a strong baseline for prompting.
    \item \textbf{DSPy} is a state-of-the-art language model programming and prompt optimization framework~\citep{khattab2024dspy}, thus we use it as the reference baseline. We instantiate DSPy's BootstrappedFewShotRandomSearch~(BFSR) optimizer with 10 candidate programs and 8 few-shot examples. This optimizer identifies demonstrations to include in the prompt as few-shot examples. This is done through generating traces of LLM inputs and outputs that individually pass the metric~(in this case, accuracy) and includes CoT reasoning. It then applies random search over subsets of up to size eight shots with these demonstrations.
\end{enumerate}

\begin{table}[!h]
\caption{\textbf{Prompt optimization for reasoning tasks.} With \textgrad, we optimize a system prompt for \texttt{gpt-3.5-turbo}~using~\gpto~as the gradient engine that provides the feedback during backpropagation.}
    \begin{tabular}{@{}lll@{}}
    \toprule
    \textbf{Dataset} & \textbf{Method}   & \textbf{Accuracy~($\%$)}\\ \midrule
    \multirow{3}{*}{Object Counting~\citep{suzgun-etal-2023-challenging, srivastava2023beyond}} & CoT (0-shot)~\citep{wei2022chain, kojima2022large} & $77.8$ \\ & DSPy (BFSR, 8 demonstrations)~\citep{khattab2024dspy} & $84.9$  \\
    & \textgrad~(instruction-only, 0 demonstrations) & $\mathbf{91.9}$  \\
    \midrule
    \multirow{3}{*}{Word Sorting~\citep{suzgun-etal-2023-challenging, srivastava2023beyond}} & CoT (0-shot)~\citep{wei2022chain, kojima2022large} & $76.7$ \\ & DSPy (BFSR, 8 demonstrations)~\citep{khattab2024dspy} & $\mathbf{79.8}$  \\
    & \textgrad~(instruction-only, 0 demonstrations) & $\mathbf{79.8}$    \\
   \midrule
    \multirow{3}{*}{GSM8k~\citep{cobbe2021gsm8k}} & CoT (0-shot)~\citep{wei2022chain, kojima2022large} & $72.9$ \\ & DSPy (BFSR, 8 demonstrations)~\citep{khattab2024dspy} & $\mathbf{81.1}$   \\
    & \textgrad~(instruction-only, 0 demonstrations) & $\mathbf{81.1}$ \\
    \bottomrule
    \end{tabular}
    \centering
    \label{tab:prompt-optimization}
\end{table}

\begin{ttcolorbox}[Example: TextGrad optimized prompt for \gptthreefive]
\noindent \textbf{\textparameter{Prompt} at initialization~(GSM8k Accuracy$=72.9\%$):}

\textit{You will answer a mathematical reasoning question. Think step by step. Always conclude the last line of your response should be of the following format: 'Answer: \$VALUE' where VALUE is a numerical value."}

\vspace{1em}

\textbf{\textparameter{Prompt} after 12 iterations with batch size 3~(GSM8k Accuracy$=81.1\%$):}

\textit{You will answer a mathematical reasoning question. Restate the problem in your own words to ensure understanding. Break down the problem into smaller steps, explaining each calculation in detail. Verify each step and re-check your calculations for accuracy. Use proper mathematical notation and maintain consistency with the context of the question. Always conclude with the final answer in the following format: 'Answer: \$VALUE' where VALUE is a numerical value.}

\end{ttcolorbox}

\textbf{Results:} Across all three tasks, \textgrad~improves the performance of the 0-shot prompt significantly. It performs similarly to DSPy~\citep{khattab2024dspy} for Word Sorting and GSM8k, and improves over DSPy by 7\% for Object Counting. While the 8 demonstrations in the context can help guide the behavior of the LLM, it can increase the cost of inference. Interestingly, the DSPy optimizer and \textgrad~make complementary adjustments---the former adds in-context demonstration examples and latter optimizes the system prompt. Adding the examples selected by DSPy to \textgrad's optimized prompt could further improve performance (for GSM8k, directly combining the demonstrations from DSPy with the instruction from TextGrad increases the accuracy to 82.1\%), suggesting that a fruitful direction is to combine both approaches.

%% file: sections/4_scientific_applications.tex
\begin{figure}[hbtp]
    \centering
    \includegraphics[width = 0.9\textwidth]{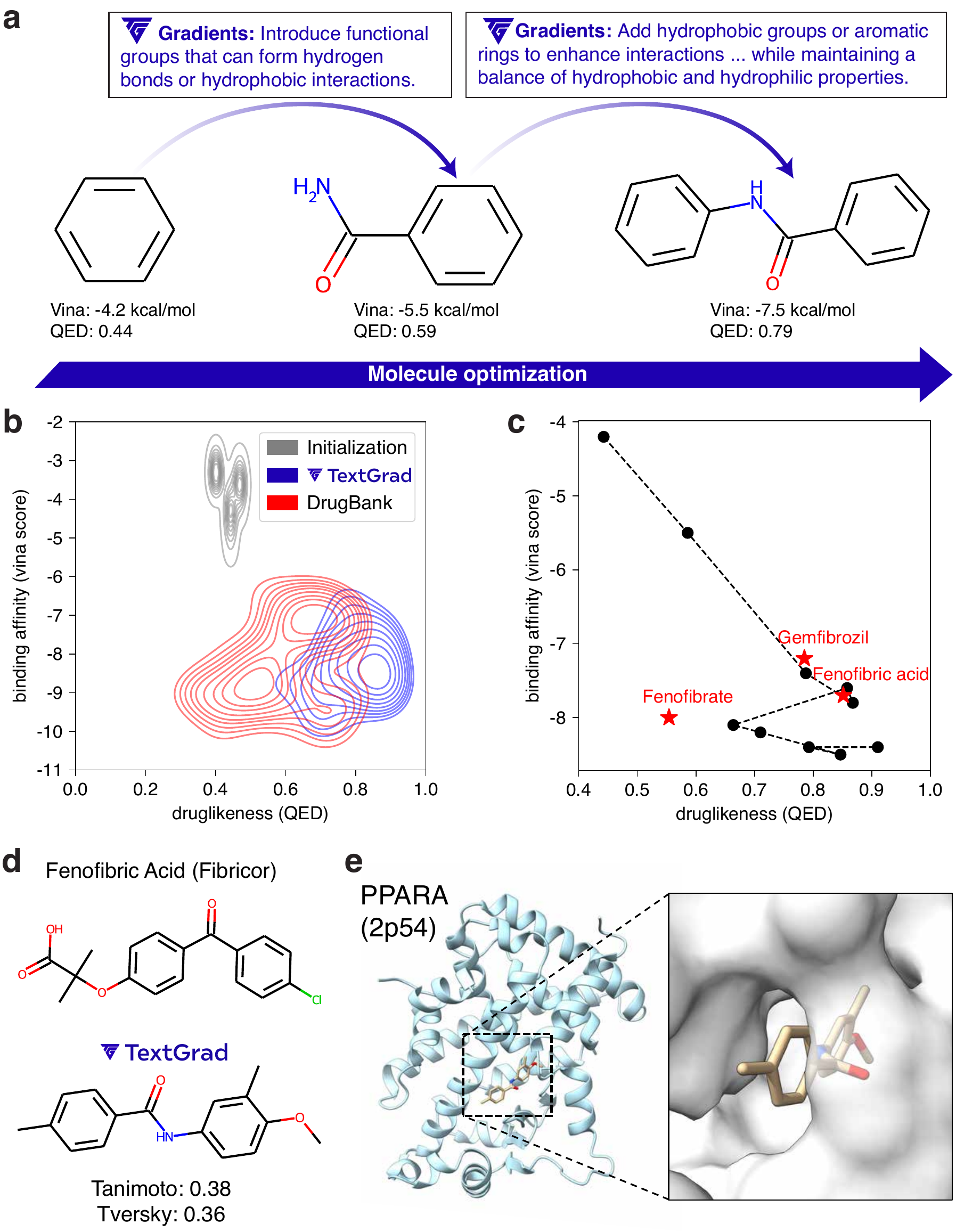}
    \caption{\textbf{Molecule optimization via Text}. \textgrad~ optimizes a starting benzene fragment to improve its druglikeness (higher QED) and binding affinity (lower vina score) to the protein receptor PPARA. The textual gradients for the first three iterations are shown in (\textbf{a}), and the performance of all ten iterations compared to clinically approved molecules targetting PPARA in (\textbf{c)}. The molecule at the final iteration has low structural similarity with its most similar clinically approved counterpart, and better QED and Vina scores \textbf{(d)} with a highly plausible pose geometry shown in \textbf{(e)}. Across 29 targets and three initial fragments, TextGrad successfully designs molecules with similar vina scores and greater QED scores than clinically approved molecules \textbf{(b)}.}
    \label{fig:mol_opt}
\end{figure}

\subsection{Molecule optimization}
\label{sec:molecule}

\textgrad~supports a variety of optimization problems, including multi-objective optimization tasks commonly found in science and engineering applications. For example, in drug discovery, researchers seek to discover or design molecules that maximize a variety of objectives with regards to synthesizability, efficacy, and safety \cite{nicolaou2013multi, hoelder2012discovery}. To demonstrate \textgrad's applications to multi-objective optimization, we apply \textgrad~ to drug molecule optimization, and show how our framework can interface with computational tools and optimize chemical structures towards simultaneously improving their binding affinity and druglikeness.

\textbf{Task:} 
A critical consideration for potential drug molecules is their \textit{binding affinity}, which represents the strength of the interactions between the molecule and its protein target. Drug designers seek molecules with high binding affinities to relevant drug targets, as they require lower and less frequent doses to achieve efficacy. This affinity can be quantified by free energy $\Delta G$, which describes the ratio of probabilities between bound and unbound ligand-receptor pairs. $\Delta G$ can be estimated using ``docking'' simulations of protein-ligand binding \cite{kontoyianni2017docking, agarwal2016overview}. In our experiments, we employ the Vina score from the Autodock Vina tool, a widely used  physics-based docking simulator \cite{trott2010autodock}. The more negative the Vina score, the greater probability that the drug will bind to its intended target. 

Potential drug molecules are also evaluated by their \textit{druglikeness}, which estimates how the molecule will behave \textit{in vivo}, with respect to solubility, permeability, metabolic stability and transporter effects. Molecules with high druglikeness are more likely to be absorbed by the body and reach their targets \cite{ursu2011understanding}. One popular metric for ``druglikeness'' is the Quantatiative Estimate of Druglikeness (QED) score, a weighted composite metric of important chemical characteristics such as molecular weight, lipophilicity, polar surface area, among others. The QED score ranges from $0$ to $1$, where $1$ indicates high druglikeness \cite{bickerton2012quantifying}.   

Though there are many more considerations for successful molecules, in our experiments, we restrict our objectives to minimizing the Vina score and maximizing the QED, due to the relative maturity of these two metrics. The competing tradeoffs  between these two metrics makes  the optimization task realistic and challenging. In particular, docking scores tend to prefer larger molecules with many functional groups that maximize interactions with a binding site \cite{bender2021practical, kontoyianni2017docking, agarwal2016overview}. In contrast, the druglikeness encourages lighter, simpler molecules that have better absorption properties \cite{bickerton2012quantifying, ursu2011understanding}. Thus,  simultaneously optimizing both objectives is non-trivial.   

\textbf{Methods:} We apply \textgrad~to drug molecule optimization by encoding molecules as SMILES strings and constructing a multi-objective loss from the Vina and QED scores. Namely, we perform instance optimization over SMILES strings, where the gradients generated by \textgrad~ with  respect to the multi-objective loss are used to update the text representing the molecule. 

\begin{align}
    \text{Evaluation} &= \textllm((\texttt{Affinity}(\textparameter{SMILES}_i, \text{target}), \texttt{Druglikeness}(\textparameter{SMILES}_i)) \\
    \textparameter{SMILES}_{i+1} &= \textcolor{RoyalPurple}{\text{TGD.step}} \left(\textparameter{SMILES}_i, \frac{\partial \text{Evaluation}}{\partial \textparameter{SMILES}_i}\right)
\end{align}

We use \gpto~ as our LLM with the prompt text found in Appendix \ref{molopt_objective_function}. At each iteration, the current molecule is evaluated by estimating its binding affinity to the target protein using the Vina score from Autodock Vina and using the QED score for druglikeness from RDKit (Appendix \ref{molopt_evaluation}). Each molecule is initialized as a small chemical fragment from a functional group. We apply \textgrad~ to all $58$ targets in the DOCKSTRING molecule evaluation benchmark \cite{garcia2022dockstring}. These $58$ targets consist of clinically relevant proteins sampled from a variety of structural classes, $29$ of which have clinically approved drugs. For each target, we optimize a starting fragment using \textgrad~ for $10$ iterations, for $3$ unique initial fragments. To evaluate our performance, we compare the characteristics of the molecules generated by \textgrad~ to clinically approved drugs for the respective protein (Appendix \ref{molopt_benchmark}). 

\textbf{Results:} For all $58$ targets, \textgrad~ consistently generates molecules with improved binding affinity and druglikeness irrespective of the initial starting fragment (Appendix \ref{molopt_initialization}). For the $29$ protein targets with clinically approved drugs, we observe that \textgrad~ generates molecules with highly competitive affinity and druglikeness when compared to clinical molecules evaluated using the same loss function (Figure \ref{fig:mol_opt} \textbf{(b)}). The resulting molecules exhibit unique structures compared to their clinical approved counterparts and existing compounds (Appendix \ref{molopt_novelty}), while maintaining similar \textit{in silico} safety profiles (Appendix \ref{molopt_admet}). While there exist alternative machine learning methods for \textit{de novo} molecular generation, \textgrad~ offers two key advantages over its counterparts: Firstly, by combining traditonal chemoinformatics tools with the the general knowledge and reasoning capabilities of LLMs, \textgrad~ produces competitive results even \textit{without a prior training set}. Secondly, \textgrad's framework of natural language gradients produce \textit{explainable} decisions, enabling researchers to understand precisely how and why a molecule's structure was constructed. Together, these two characteristics invoke a promising future for the role of AI agents in scientific discovery.

\begin{figure}[hbtp]
    \centering
    \includegraphics[width = \textwidth]{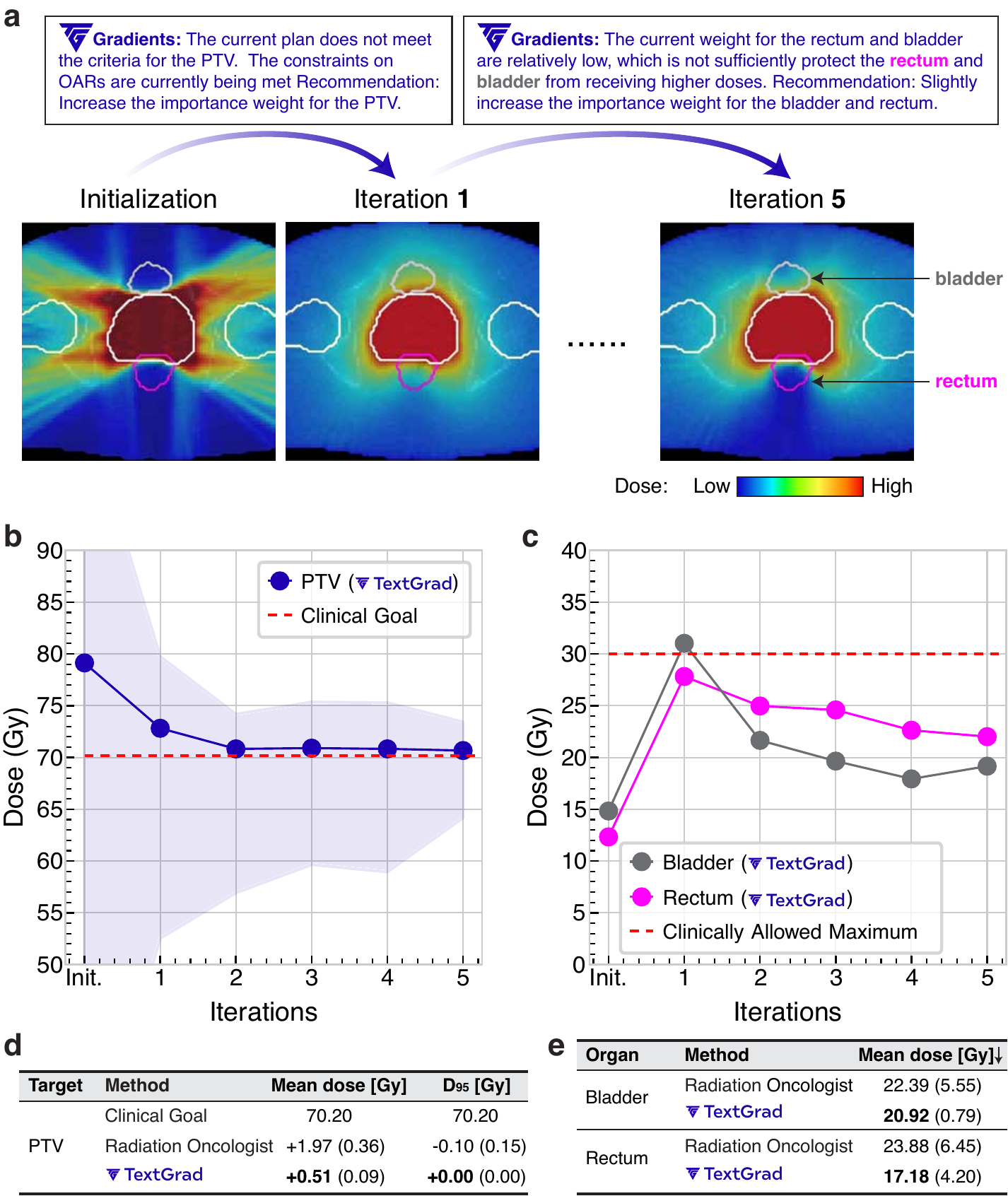}
    \caption{\textbf{Radiotherapy treatment plan optimization.} Visualization of backpropagated textual gradients adjusting importance weights of planning target volumes (PTVs) and organs at risk (OARs) to balance tumor targeting and protection of OARs. Images in \textbf{(a)} show plan evolution from initialization to iteration 5. \textgrad~iteratively improves the mean dose and reduces the dose variance to the PTV, achieving the clinical goal over multiple iterations \textbf{(b)}. \textgrad~keeps the exposure for the bladder and rectum below clinically allowed maximums \textbf{(c)}. \textgrad~optimized plans have better dose metrics (mean dose and $D_{95}$) for PTV than clinically optimized plans~\textbf{(d)}, and lower doses on bladder and rectum, indicating better protection of OARs \textbf{(e)}.
}
    \label{fig:treatment_plan}
\end{figure}

\subsection{Radiotherapy treatment plan optimization}
\label{sec:treatment-plan}

Radiation therapy, also known as radiotherapy, is a cancer treatment that uses beams of intense energy, such as X-rays, to kill cancer cells. Before treatment begins, a radiotherapy team, including radiation oncologists and planners, collaborates to design an effective treatment plan. This involves determining the necessary dose of radiotherapy and pinpointing the exact locations that need treatment.

Radiotherapy treatment planning can be formulated as a two-loop optimization problem. The inner loop, known as inverse planning, includes processes such as influence map optimization and direct aperture optimization~\cite{khan2021khan}. This optimization problem is typically a constrained one, solved by a numerical optimizer, aiming to minimize a weighted cost function balancing multiple conflicting objectives~\cite{webb2003physical}. These objectives include delivering the prescribed dose to the planning target volume (PTV), which encompasses the tumor and an additional margin to account for uncertainties in planning or treatment delivery, while protecting critical normal tissues, known as organs at risk (OARs), from receiving unsafe doses.

The main challenge in treatment planning is translating overall clinical goals into weighted objective functions and dose constraints that yield an acceptable plan~\cite{khan2021khan}. Human planners often use a trial-and-error approach, iteratively adjusting optimization hyperparameters based on the results of the optimization process until the plans meet clinical requirements~\cite{khan2021khan}. These hyperparameters include the weights assigned to PTVs, organs, and other tissues in the objective function. This process can be subjective, influenced by the planner's experience and the available time, and involves repeatedly using computationally expensive optimization algorithms over many iterations. This makes the process inefficient, time-consuming, and costly~\cite{husseinAutomationIntensityModulated2018a}.

\paragraph{Method.} We apply \textgrad~to perform the outer loop optimization, i.e. hyperparameter optimization for the inner loop numerical optimizer. Instance optimization is performed with \gpto~over the hyperparameters represented as a string: $\theta=$ ``weight for PTV: [PTV WEIGHT], weight for bladder: [BLADDER WEIGHT], weight for rectum: [RECTUM WEIGHT], weight for femoral heads: [FH WEIGHT], weight for body: [BODY WEIGHT]''. When hyperparameters are provided, we obtain the treatment plan by adopting a numerical optimizer and constructing a loss as the mismatch between the current plan and the clinical objectives. Specifically, to compute the gradient, we first solve the inner optimization loop using a numerical optimizer matRad\cite{wieser2017development} to obtain the corresponding treatment plan $P(\theta) = \texttt{matRad}(\theta)$. The loss is computed on the treatment plan $P$ and the clinical goals $g$ using an LLM with prompts provided in Section~\ref{tp_prompts}.
\[
\mathcal{L} = \texttt{LLM}(P(\theta),g)
\]
and the new hyperparameters are obtained by a TextGrad descent step 
\[
\theta_{\textrm{new}} = \textcolor{NavyBlue}{\text{TGD.step}}\left(\theta, \frac{\partial \mathcal{L}}{\partial \theta}\right).
\]
To further improve LLM's capability to understand the relationship between hyperparameters $\theta$ and the resulting plan $P$ from matRad, a set of paired plans and their corresponding hyperparameters $\{(P_i, \theta_i)\}_{i=1}^N$ is provided as context for LLMs during the $\textcolor{NavyBlue}{\text{TGD.step}}$. Therefore,
\[
\theta_{\textrm{new}} = \textcolor{NavyBlue}{\text{TGD.step}}\left(\theta, \frac{\partial \mathcal{L}}{\partial \theta}\middle|\{(P_i, \theta_i)\}_{i=1}^N\right).
\]
\paragraph{Evaluation metrics.} 
To evaluate a treatment plan, we adopt several commonly used dose metrics as a plan cannot be evaluated using a single metric. We consider the mean dose delivered to the target/organ volume, as well as $D_{q}$, which denotes the minimum dose received by $q$\% of the target/organ volume.    

\paragraph{Results.}
The gradients generated by \textgrad~provide meaningful guidance to improve the hyperparameters. As illustrated in~Figure~\ref{fig:treatment_plan}, when there is dose spillage outside the Planning Target Volume~(PTV), the gradient suggests an increase in the importance weight for the PTV. This adjustment results in a more uniform and confined dose for the PTV. However, this can lead to insufficient protection of the bladder and rectum as their relative weights are reduced. Therefore, in the following step, the gradients suggest slightly increasing the weights for the bladder and rectum, resulting in better protection for these organs. We compared \textgrad~optimized plans with the clinical plans used to treat five prostate cancer patients.  In Figure~\ref{fig:treatment_plan}~(c), we assess TextGrad's capabilities in achieving clinical goals for the PTV region. TextGrad outperforms the clinical plans across all metrics, achieving a higher mean dose, and a $D_{95}$
that exactly matches the prescribed dose. In Figure~\ref{fig:treatment_plan}~(d), we focus on the sparing of healthy organs. TextGrad-optimized plans achieve lower mean doses for these healthy organs, suggesting better organ sparing than the human-optimized plans. We report the averages across five plans and with standard deviation included in the bracket.

%% file: sections/5_related_works.tex
One related thread of work investigated the problem of prompt optimization.  Practitioners demonstrated that prompt engineering strategies such as intelligently picking few-shot examples and in-context learning, CoT, ensembles can significantly boost performance of LLMs~\citep{nori2023can}. To automate this process, white-box methods that leverage numerical gradients were developed to optimize prompts~\citep{shin-etal-2020-autoprompt, jia2022visual, li-liang-2021-prefix, chen2022knowprompt}, however, these methods cannot be used with closed-source models as they require access to model parameters. Various works investigated using LLMs as prompt optimizers~\citep{zhou2022large, ye2023prompt, pryzant-etal-2023-automatic}.

Under prompt optimization, there are two works closest to our philosophy that have been our inspirations. First, DSPy~\citep{khattab2024dspy, khattab2022demonstrate, singhvi2023dspy} pioneered the idea of viewing complex LLM-based systems as programs with potentially many layers, and proposes ways to build and optimize them in a programmatic fashion. The framework is extensive, with results improving LLM performance in various question answering, reasoning, and prompt optimization tasks. Our work takes a different perspective that backpropagation and its extensions can be a general and powerful framework to optimize the new generation of AI systems, and perform multiple tasks outside of prompt optimization. In particular, we treat not only instructions or demonstrations as variables to optimize, but also the instances we care about themselves --- such as molecules, treatment plans, code snippets, and so on. Second, greatly inspiring to us, Prompt Optimization with Textual Gradients~(ProTeGi)~\citep{pryzant-etal-2023-automatic} defines the Textual Gradients in the context of prompt optimization, where gradients are natural language feedback from LLMs given to the mistakes made during the task. While ProTeGi is built on the textual gradient analogy, we expand this analogy more broadly to automatic differentiation, and going substantially beyond prompt optimization tasks. In particular, both DSPy and ProTeGi focused on prompt optimization, while a significant advance of \textgrad, as demonstrated through our diverse applications, is in instance optimization.

More generally, there is an emerging line of work built on the high-level idea of using LLMs as critics or optimizers~\citep{pryzant-etal-2023-automatic, yang2024large, reflexion, madaan2024self, song2024position, khattab2024dspy, liu2024large, ye2023prompt, wang2024hypothesis, zhou2022large, gao-etal-2023-rarr, chen2024teaching, shypula2024learning}. While many of these earlier frameworks demonstrated the utility of LLMs as optimizers, we propose a single and general framework that was tested successfully in a variety of applications. Within this framework, we can reason about optimizing chains or stacks of LLMs~\citep{Chase2022LangChain, liu2022llamaindex, schick2024toolformer}: we propagate natural language feedback. Similarly, once viewed as a general-purpose optimization engine, we can formulate many relevant problems instantiated as a few lines of code in our framework, such as test-time training~\citep{sun2023learning, sun2020test} or self-refinement of solutions and self-improvement~\citep{madaan2024self, press-etal-2023-measuring, reflexion, zelikman2022star, yao2023react, huang-etal-2023-large, yang-etal-2022-re3, xie2023selfevaluation, paul-etal-2024-refiner, zhao2024expel, le2024codechain}. Building on the optimization analogy, we already transferred several analogies from the traditional optimization literature such as momentum~\citep{polyak1964some} through using earlier iterations in the context, use of batch optimization~\citep{robbins1951stochastic}, constrained optimization ~\citep{boyd2004convex} using natural language constraints, and so on. Our work opens up a large space to design the new generation of optimization algorithms, all within the same framework.

%% file: sections/6_discussion.tex
TextGrad is built on three key principles: i) It is a general and performant framework that is not hand-crafted for a specific application domain, ii) It is easy-to-use, mirroring PyTorch abstractions thus allowing knowledge transfer, iii) It is fully open-source. 
Through \textgrad, we obtained state-of-the-art results in code optimization and PhD-level question answering, optimized prompts, and provided proof-of-concept results in scientific applications such as developing molecules and optimizing treatment plans.

While we took a first step, there are various limitations that motivate future work to realize the potential of automatic differentiation frameworks powered by LLMs. 
First, while we demonstrated the potential of backpropagating text feedback, there are many applications our framework can be extended to. We hope \textgrad~can be used to accelerate iterative processes in scientific discovery and increase the productivity of engineering efforts. For instance, to allow for this, we hope to extend the operations in our computation graphs to include more components used in practical LLM applications, such as for tool use~\citep{schick2024toolformer} or retrieval-augmented generation systems~\citep{lewis2020retrieval}.
Second, the automatic differentiation analogy enables a large design space for algorithms. We believe there are many fruitful connections to be drawn between numerical optimization, automatic differentiation, and \textgrad. In particular, increasing the stability of the optimization using variance reduction techniques~\citep{robert1995simulation}, adaptive gradients~\citep{KingBa15}, or self-verification using LLMs~\citep{lightman2024lets} are interesting connections. Meta learning approaches~\citep{schmidhuber1987evolutionary, thrun1998learning, finn2017model} to optimize the TextGrad framework using methods such as TextGrad itself is also an intriguing direction of future work.

Finally, while we conducted proof-of-concept applications of \textgrad~to design new molecules and treatment plans with \textit{in silico} validations, the ultimate test requires experimental and clinical assessments, which are outside of the scope of this paper. 

As the paradigm of AI shifts from training individual models to optimizing compound systems involving multiple interacting LLM components and tools, we need a new generation of automated optimizers. \textgrad~combines the reasoning power of LLMs with the decomposable efficiency of backpropation to create a general framework to optimize AI systems.

%% file: appendix/main.tex
\input{appendix/textgrad_operations}

\section{Code Optimization}
\label{appendix:code-optimization}
\input{appendix/code_optimization}

\section{Solution Optimization}
\label{appendix:solution-optimization}
\input{appendix/solution_optimization}

\section{Prompt Optimization}
\label{appendix:prompt-optimization}
\input{appendix/prompt_optimization}

\section{Molecule Optimization}
\label{appendix:molecule-optimization}
\input{appendix/molecule_optimization}

\section{Treatment Plan Optimization}
\label{appendix:treatment-plan-optimization}
\input{appendix/treatment_plan}

%% file: appendix/textgrad_operations.tex
\section{\textgrad~Details}
\label{appendix:textgrad-operations}

\subsection{Variables}
\texttt{Variable}s are the nodes in the computation graph. Below are the most important attributes of \texttt{Variable}s:
\begin{enumerate}
    \item \textbf{Value} is the unstructured data that the variable contains. Throughout this manuscript, all values are text data.
    \item \textbf{Role description} is an informative string that describes the role of the variable in the computation graph. We use these roles to let the user inject knowledge into the graph and guide the optimization behavior. More information is described below.
    \item \textbf{Gradients} are the natural language feedback provided by the LLMs during the backward pass. These describe the changes to make the variable so that the downstream loss can be improved. For an example backward operation that populates gradients, please read Section~\ref{appendix:llm-op}.
    \item \textbf{Predecessors} are the set of variables that are used to generate a given variable. For instance, if we are giving an instruction to an LLM and getting a response, the instruction would be the predecessor of the response. During the backward pass, the gradients on the successor are passed to its predecessors, to provide guidance around how to improve the downstream objective.
    \item \textbf{Requires grad} indicates whether or not the gradients will be populated during the backward pass, analogous to PyTorch. For instance, if the user does not wish to compute gradients for a question, then simply write \texttt{Variable(value=question, requires\_grad=False, \ldots)} to indicate this.
\end{enumerate}

\textbf{Role Description:} In \textgrad, each variable has a \textit{role description}. In particular, this is a string that describes the role of the variable in the computation graph, such as \textit{system prompt to the language model} or \textit{prediction by the language model}. While sometimes populated automatically, in general role descriptions are one of the primary ways to inject user knowledge into the optimization process. 

Empirically, we find that role descriptions can significantly steer the optimization process. For instance, setting the role of a prediction to be $\textit{the final numerical answer to the language model}$ guides the Textual Gradient Descent optimizer, that prompts a language model to update the value of the variable using the feedback, to update the variable such that it is only a numerical value. In comparison, a role description such as $\textit{the reasoning for the solution and the final prediction}$ guides the optimizer to produce the reasoning along with the final prediction.

\noindent Here is an example usage:

\noindent\begin{minipage}[!h]{\textwidth}
    \centering
    \begin{lstlisting}[language=Python, caption=An example usage of a role description.]
    import textgrad as tg
    
    system_prompt = tg.Variable("You will be given a question and think step-by-step.", requires_grad=True, role_description="system prompt to the language model that will be reused across queries")

    model = tg.BlackBoxLLM(system_prompt=system_prompt)\end{lstlisting}
    \label{fig:role_description}
\end{minipage}

\subsection{Backpropagation}
\label{appendix:backprop}

The backpropagation algorithm in \textgrad~ mirrors existing autograd frameworks for deep learning. See Algorithm~\ref{appendix:backprop-algorithm} for the pseudocode.

\begin{algorithm}
\caption{Backpropagation in \textgrad}
\label{appendix:backprop-algorithm}
\begin{algorithmic}[1]
\STATE \textbf{Input:} Variables $v \in \mathcal{V}$ in a graph, Loss variable $\mathcal{L}$, Backward Engine~(LLM) $\mathcal{M}$ that will provide textual gradients
\STATE \COMMENTT{Initializing gradients}
\FOR{each $v \in \mathcal{V}$}
    \STATE $v.\text{gradients} = \emptyset$
\ENDFOR
\STATE \COMMENTT{Topological Sorting}
\STATE $Q \gets \text{TopologicalSort}(G)$
\STATE \COMMENTT{Backpropagation}
\FOR{$v$ in $Q$}
    \STATE \COMMENTT{Populate gradients in predecessors}
    \FOR{each $u \in \text{PredecessorsOf}(v)$}
        \STATE \COMMENTT{Here, we are omitting subscript $v$ in $f$. Semantically, $f$ is the function that generates $v$, and $\textbackwardchain{f}$ is the backward operation for that function.}
        \STATE \COMMENTT{Semantically, this provides feedback to the variable $u$, given how $v$ is produced, and the feedback we already collected for $v$.}
        \STATE $u.\text{gradients.add}\left (\mathlarger{\textbackwardchain{f}}\left(u, v,  \frac{\partial \mathcal{L}}{\partial v}\right)\right )$
    \ENDFOR
\ENDFOR
\end{algorithmic}
\end{algorithm}

\subsection{Functions}
\label{appendix:functions}
\textgrad~offers several operations where both the forward and backward computations are defined -- as such, these operations are composable. 
The abstract \texttt{textgrad.autograd.Function} class has two methods: \texttt{forward} and \texttt{backward}, mirroring the PyTorch syntax. Each function has to define both of these methods. Below, we describe a couple of the most used operations in this paper.

\paragraph{\texttt{LLMCall} Function:}
\label{appendix:llm-op}
Currently, the most crucial operation in \textgrad~is the call to language models. 

\textbf{Forward mode.} The forward mode is simple: We make a call to an LLM, through an API or through the local machine. When a call is made, all the input variables are registered as the predecessors of the response from the LLM. For instance, if we ask a question to an LLM using an \texttt{instruction} and a \texttt{question} variable, the \texttt{response} variable's predecessors will be \texttt{[instruction, question]}. When doing the backward pass, the gradients on the \texttt{response} will be backpropagated to \texttt{question} and \texttt{instruction}.

\textbf{Backward mode.} To ensure the backward function runs smoothly and generally, we add the following glossary to the system prompt:

\begin{apxtcolorbox}[Glossary for Backward Mode of the \texttt{LLMCall} function]
\#\#\# Glossary of tags that will be sent to you:

\# - <LM\_SYSTEM\_PROMPT>: The system prompt for the language model.

\# - <LM\_INPUT>: The input to the language model.

\# - <LM\_OUTPUT>: The output of the language model.

\# - <OBJECTIVE\_FUNCTION>: The objective of the optimization task.

\# - <VARIABLE>: Specifies the span of the variable.

\# - <ROLE>: The role description of the variable.
\end{apxtcolorbox}

Using these tags, the LLM is made aware of the objective, the role and the value of the variable to give feedback to, and the full conversation in the forward mode. 

This glossary is provided in the system prompt to the backward engine LLM:

\begin{apxtcolorbox}[System prompt for the backward mode of the \texttt{LLMCall} function]
You are part of an optimization system that improves a given text (i.e. the variable). You are the gradient (feedback) engine. Your only responsibility is to give intelligent and creative feedback and constructive criticism to variables, given an objective specified in <OBJECTIVE\_FUNCTION\> </OBJECTIVE\_FUNCTION> tags. The variables may be solutions to problems, prompts to language models, code, or any other text-based variable. Pay attention to the role description of the variable, and the context in which it is used. You should assume that the variable will be used in a similar context in the future. Only provide strategies, explanations, and methods to change in the variable. DO NOT propose a new version of the variable, that will be the job of the optimizer. Your only job is to send feedback and criticism (compute 'gradients'). For instance, feedback can be in the form of 'Since language models have the X failure mode...', 'Adding X can fix this error because...', 'Removing X can improve the objective function because...', 'Changing X to Y would fix the mistake ...', that gets at the downstream objective.

If a variable is already working well (e.g. the objective function is perfect, an evaluation shows the response is accurate), you should not give feedback.

\{GLOSSARY\}
\end{apxtcolorbox}

Most of this setup aims to ensure that the user does not have to modify the gradient computation. All of our experiments in the diverse set of applications are done with the same backward mode.

For instance, an example prompt for the gradient computation looks like the following:

\begin{apxtcolorbox}[An example backward mode prompt for the \texttt{LLMCall} function]
\footnotesize
You will give feedback to a variable with the following role: <ROLE> system prompt to a language model </ROLE>. Here is a conversation with a language model (LM):

<LM\_SYSTEM\_PROMPT> You will answer a reasoning question. Think step by step. Always conclude the last line of your response should be of the following format: 'Answer: \$VALUE' where VALUE is a numerical value. </LM\_SYSTEM\_PROMPT>

<LM\_INPUT> I have a chicken, a frog, a mouse, a cat, two pigs, and two rabbits. How many animals do I have? </LM\_INPUT>

<LM\_OUTPUT> To find the total number of animals, we simply add up the number of each type of animal: 1 chicken + 1 frog + 1 mouse + 1 cat + 2 pigs + 2 rabbits = 8 animals.

Answer: 8 </LM\_OUTPUT>

This conversation is part of a larger system. The <LM\_OUTPUT> was later used as response from the language model.

<OBJECTIVE\_FUNCTION>Your goal is to give feedback to the variable to address the following feedback on the LM\_OUTPUT: To improve the runtime of the string-based function that checks if the prediction is correct, consider the following feedback:

1. **Simplify the Response**: For example, instead of "To find the total number of animals, we simply add (...) + 2 pigs + 2 rabbits = 8 animals. Answer: 8", a more concise response like "Total animals: 8" would be more efficient.

\ldots

By implementing these strategies, the response from the language model can be optimized to improve the runtime of the string-based function that checks if the prediction is correct. </OBJECTIVE\_FUNCTION>

We are interested in giving feedback to the system prompt to a language model. Specifically, give feedback to the following span of text:

<VARIABLE> You will answer a reasoning question. Think step by step. (...) following format: 'Answer: \$VALUE' where VALUE is a numerical value. </VARIABLE>

Given the above history, describe how the system prompt to a language model could be improved to improve the <OBJECTIVE\_FUNCTION>. Be very creative, critical, and intelligent.
\end{apxtcolorbox}

\paragraph{Addition Operation}
In numerical optimization, we have the following:
\begin{align}
    z & = x+y  \\
    w & = t+z \\
    \frac{\partial w}{\partial x} & = \frac{\partial w}{\partial z}\frac{\partial z}{\partial x} = \frac{\partial w}{\partial z}
\end{align}

In numerical derivatives, due to the linearity of addition, we have:
\begin{equation}
\frac{\partial \sum_i^N L_i}{\partial {x}} = \sum_{i}^N \frac{\partial L_i}{\partial {x}}.
\end{equation}
In particular, the backward function for the addition operation passes the gradients on the output of the addition operation to its inputs.

In particular, the backward function for the addition operation passes the gradients on the output of the addition operation to its inputs.

\begin{equation}
    \textttgrad{tg.sum(}L_1, \ldots, L_N\textttgrad{)} = "\{L_1\} \text{\textbackslash} n \{L_2\} \text{\textbackslash} n \ldots \{L_N\}"
\end{equation}

Similarly, in \textgrad, we have the $\textttgrad{tg.sum}$ operation, that lets the gradients~(feedback) on the output variable pass to the input variables. 

\begin{equation}
    \frac{\partial \textttgrad{tg.sum(}L_1, \ldots, L_N\textttgrad{)}}{\partial x} = \textttgrad{tg.sum(}\frac{\partial L_1}{\partial x}, \ldots, \frac{\partial L_N}{\partial x}\textttgrad{)} =  "\{\frac{\partial L_1}{\partial x}\} \text{\textbackslash} n \ldots \{\frac{\partial L_N}{\partial x}\}"
\end{equation}

where we use $+$ to indicate concatenation.

\textbf{Use Case: } One canonical use case for the addition operation is the batch optimization case. In particular, we implement minibatch gradient descent when performing prompt optimization~(Section~\ref{sec:prompt-optimization}).

\subsection{Textual Gradient Descent Implementation}
Similar to backward computations, we strive to preserve generality in the implementation of TGD. In particular, we use the same glossary of tags provided above to inject information to the optimization process.

The current system prompt to the optimizer call is the following:

\begin{apxtcolorbox}[System prompt for the \texttt{TextualGradientDescent} optimizer.]
You are part of an optimization system that improves text (i.e., variable). You will be asked to creatively and critically improve prompts, solutions to problems, code, or any other text-based variable. You will receive some feedback, and use the feedback to improve the variable. The feedback may be noisy, identify what is important and what is correct. Pay attention to the role description of the variable, and the context in which it is used. This is very important: You MUST give your response by sending the improved variable between <IMPROVED\_VARIABLE> {{improved variable}} </IMPROVED\_VARIABLE> tags. The text you send between the tags will directly replace the variable.

\{GLOSSARY\}
\end{apxtcolorbox}

Below is an example prompt to the optimizer:

\begin{apxtcolorbox}[An example prompt to the optimizer]
\footnotesize
Here is the role of the variable you will improve: <ROLE>system prompt to the language model</ROLE>.

The variable is the text within the following span: <VARIABLE> You will answer a reasoning question. Think step by step. (...) following format: 'Answer: \$VALUE' where VALUE is a numerical value. </VARIABLE>

Here is the context and feedback we got for the variable:

<CONTEXT>

Here is a conversation:

<CONVERSATION><LM\_SYSTEM\_PROMPT> You will answer a reasoning question. Think step by step. The last line of your response should be of the following format: 'Answer: \$VALUE' where VALUE is a numerical value. </LM\_SYSTEM\_PROMPT>

<LM\_INPUT> I have a carrot, an onion, a stalk of celery, three yams, a garlic, a head of broccoli, and a potato. How many vegetables do I have? </LM\_INPUT>

<LM\_OUTPUT> To determine the number of vegetables you have, we need to count each individual vegetable item. 

1. Carrot
2. Onion
3. Celery
4. Yams (3 yams)
5. Garlic
6. Broccoli
7. Potato

Counting all the individual vegetable items, we have a total of 7 vegetables.

Therefore, you have 7 vegetables.

Answer: 7 </LM\_OUTPUT>

</CONVERSATION>

This conversation is potentially part of a larger system. The output is used as response from the language model

Here is the feedback we got for system prompt to the language model in the conversation:

<FEEDBACK>To improve the structured system prompt for the language model, consider the following feedback:

1. **Clarify the Calculation Process**:
   - **Current Issue**: The prompt does not explicitly instruct the model to outline each step of its calculation process.
   - **Improvement**: Add a directive that requires the model to explicitly state each quantity and the sum. For example, "Clearly outline each step of your calculation process, stating each quantity and the sum."

2. **Verify Intermediate Steps**:
   - **Current Issue**: The prompt does not instruct the model to verify each intermediate step.
   - **Improvement**: Include a directive for the model to verify each step against known correct values. For example, "Verify each intermediate step in your calculation to ensure accuracy."

</FEEDBACK>

</CONTEXT>

Improve the variable (system prompt to the language model) using the feedback provided in <FEEDBACK> tags.
Send the improved variable in the following format:

<IMPROVED\_VARIABLE>{the improved variable}</IMPROVED\_VARIABLE>

Send ONLY the improved variable between the <IMPROVED\_VARIABLE> tags, and nothing else.

\end{apxtcolorbox}

\section{Optimizer Extensions}
\label{appendix:optimizer-extensions}

\subsection*{Batch Optimization}
In batch optimization, we use the \texttt{tg.sum} function described above. In particular, gradients propagating from multiple instances are concatenated together, thus the optimizer sees all of the feedback to a variable coming from multiple sources.

The syntax is as simple as the following:

\noindent\begin{minipage}[!h]{\textwidth}
    \centering
    \begin{lstlisting}[language=Python, caption=An example use for batch optimization for question answering.]
losses = [loss_fn(answer, model(question)) for question, answer in batch]
total_loss = tg.sum(losses)
total_loss.backward()\end{lstlisting}
    \label{fig:batch_optimization}
\end{minipage}

\subsection*{Constrained Optimization with Natural Language Constraints}

In \textgrad~it is possible use constraints when optimizing variables. These constraints are all defined as natural language descriptions. For example, one can prompt optimizer to update the variable but to conclude its response with an answer during the update: 

\begin{apxtcolorbox}[Example constrained optimization prompt]
You must follow the following constraints:

<CONSTRAINTS>\textit{`The last line of your response should be of the following format: 'Answer: \$LETTER' where LETTER is one of ABCD.''}</CONSTRAINTS>
\end{apxtcolorbox}

In general, the constraint post-fix is appended to the optimizer's prompt, where the constraints are written within the \texttt{<CONSTRAINTS> \{constraint text\} </CONSTRAINTS>} tags.

In code, the user can simply pass in the constraints to the TGD optimizer:

\noindent\begin{minipage}[!h]{\textwidth}
    \centering
    \begin{lstlisting}[language=Python, caption=An example use for constraints when updating the solution to a problem.]
constraints = ["The last line of your response should be of the following format: 'Answer: \$LETTER' where LETTER is one of ABCD."]
optimizer = TextualGradientDescent(parameters=[solution], constraints=constraints)\end{lstlisting}
    \label{fig:tgd_constraints}
\end{minipage}

\subsection*{Momentum}

\textgrad~supports the use of Momentum in the Textual Gradient Descent. In standard SGD momentum uses a linear combination of past gradients and the most recent one to define a new gradient to update a variable. Similarly, \textgrad~keeps track of past iterations of the variable. This postfix is appended to the prompt for the optimizer.

\noindent\begin{minipage}[!h]{\textwidth}
    \centering
    \begin{lstlisting}[language=Python, caption=How to enable momentum using 3 previous steps in the \texttt{TextualGradientDescent} optimizer.]
optimizer = TextualGradientDescent(parameters=[solution], momentum_window=3)\end{lstlisting}
    \label{fig:tgd_momentum}
\end{minipage}

\begin{apxtcolorbox}[Momentum prompt]
Here are the past iterations of this variable:

<PAST\_ITERATIONS>\{past\_values\}</PAST\_ITERATIONS>
\end{apxtcolorbox}

\subsection*{In-Context Examples}
In-context examples can be utilized to improve textual gradients and update variables effectively. These examples serve as references to illustrate the characteristics of optimized variables. When in-context examples are applied, \textgrad~adopts the following prompt to incorporate them:  

\begin{apxtcolorbox}[In-context examples prompt]
You must base on the following examples when modifying the \{role\_description\}:

<EXAMPLES>\{in\_context\_examples\}</EXAMPLES>"
\end{apxtcolorbox}

By leveraging these examples, \textgrad~can better understand and implement the properties of optimized variables, enhancing the overall optimization process

%% file: appendix/code_optimization.tex
\subsection{Methodology}

\textbf{Baseline Details.}
We re-run Reflexion~\cite{reflexion} using the code by the authors currently available online. We had to make minor changes to ensure it ran correctly in our setting but we also contacted the original authors and ask feedback on our edits to ensure our evaluation was consistent. 
Moreover, we had to re-extract the LeetCodeHard~\cite{reflexion} dataset using the authors' pipeline; this means that it is likely that the dataset we are using in this manuscript is not the same dataset that was used in the Reflexion paper. While this dataset contains a set of simple tests to check if the code works as expected, the real evaluation in this context is given by passing the tests on the LeetCode platform. 

Reflexion is run with a \textit{one shot prompt} that is meant to instruct the model on how to provide feedback. The pipeline run by Reflexion is as follows: given in input a prompt to generate code, a language model generates a first solution. If this solution passes the local tests, it is then submitted to the LeetCode platform to be evaluated on harder tests. However, if the solution does not pass the tests, we ask for feedback through Reflexion and optimize the code. Once again, if the new solution passes the local tests, we submit it to the LeetCode platform. We do this optimization for 5 iterations.

We ran the experiment 5 times with 5 different seeds and we averaged the results. At each iteration of optimization, \textgrad makes 1 call to \gpto~to evaluate the test time loss, 1 call to collect gradients, and 1 call to update the code snippet. The number of coding problems in LeetCodeHard is 39.

\begin{apxtcolorbox}[Example Query for LeetCode Hard]
\footnotesize
def minimumTime(grid: List[List[int]]) -> int:
    
    """
    
    You are given a `m x n` matrix `grid` consisting of non-negative integers where `grid[row][col]` represents the minimum time required to be able to visit the cell `(row, col)`, which means you can visit the cell `(row, col)` only when the time you visit it is greater than or equal to `grid[row][col]`.
    
    You are standing in the top-left cell of the matrix in the `0th` second, and you must move to any adjacent cell in the four directions: up, down, left, and right. Each move you make takes 1 second.
    
    Return the minimum time required in which you can visit the bottom-right cell of the matrix. If you cannot visit the bottom-right cell, then return `-1`.
    
    Example 1:
    
    Input: grid = [[0,1,3,2],[5,1,2,5],[4,3,8,6]]
    
    Output: 7
    
    Explanation: One of the paths that we can take is the following:
    
    - at t = 0, we are on the cell (0,0).
    
    - at t = 1, we move to the cell (0,1). It is possible because grid[0][1] <= 1.
    
    - at t = 2, we move to the cell (1,1). It is possible because grid[1][1] <= 2.
    
    - at t = 3, we move to the cell (1,2). It is possible because grid[1][2] <= 3.
    
    - at t = 4, we move to the cell (1,1). It is possible because grid[1][1] <= 4.
    
    - at t = 5, we move to the cell (1,2). It is possible because grid[1][2] <= 5.
    
    - at t = 6, we move to the cell (1,3). It is possible because grid[1][3] <= 6.
    
    - at t = 7, we move to the cell (2,3). It is possible because grid[2][3] <= 7.
    
    The final time is 7. It can be shown that it is the minimum time possible.
    
    Example 2:
    
    Input: grid = [[0,2,4],[3,2,1],[1,0,4]]
    
    Output: -1
    
    Explanation: There is no path from the top left to the bottom-right cell.
    
    Constraints:
    
    * `m == grid.length`
    
    * `n == grid[i].length`
    
    * `2 <= m, n <= 1000`
    
    * `4 <= m * n <= 105`
    
    * `0 <= grid[i][j] <= 105`
    
    * `grid[0][0] == 0`
    
    """
\end{apxtcolorbox}

%% file: appendix/solution_optimization.tex
\subsection{Methodology}
\label{appendix:solution-optimization-methodology}

For the CoT 0-shot prediction, we use the question template and system prompt released with GPT-4o in the \href{https://github.com/openai/simple-evals/}{simple-evals} repository. In particular, to closely match their evaluations, we use the ChatGPT system prompt: \textit{You are ChatGPT, a large language model trained by OpenAI, based on the GPT-4 architecture. \textbackslash n Knowledge cutoff: 2023-12 \textbackslash n Current date: 2024-04-01"
}. Further, we use the following template:

\begin{apxtcolorbox}[Multiple Choice Question Answering Template]
    Answer the following multiple choice question. The last line of your response should be of the following format: 'Answer: \$LETTER' (without quotes) where LETTER is one of ABCD. Think step by step before answering.

\{Question\}

A) \{A\}

B) \{B\}

C) \{C\}

D) \{D\}
\end{apxtcolorbox}

During optimization, we provide the constraint to the optimizer that the prediction should conclude with an answer, following the simple-evals repository, through the following constraint description: \textit{The last line of your response should be of the following format: 'Answer: \$LETTER' (without quotes) where LETTER is one of ABCD.}.

\textbf{Evaluation:} Similarly, using the practice in the \href{https://github.com/openai/simple-evals/}{simple-evals} repository, we perform string matching to find the final answer, which is one of the letters ABCD, and compare it to the ground truth answer. GPQA Diamond subset contains $198$ questions. MMLU Machine Learning subset contains $112$ questions, and College Physics subset contains $92$ questions. At each iteration of optimization, we make 1 call to \gpto~to evaluate the test time loss, 1 call to collect gradients, and 1 call to update the solution.

\subsection{Prompts}
\label{appendix:solution-optimization-prompts}
The loss function for this task looks like the following:

\begin{apxtcolorbox}[Solution Refinement Objective]
Below is a multi-choice question and a prediction. You are a critical and creative scientist. Your job is to investigate the prediction. Critically go through reasoning steps, and see if there is a reason why the prediction could be incorrect.

Use the Janusian Process, think about whether alternative answers could be true.

Question: $\{\text{Question}\}$

Answer by the language model: $\{\textparameter{Solution}\}$
\end{apxtcolorbox}

\begin{apxtcolorbox}[Example Query for GPQA Diamond]
What is the concentration of calcium ions in a solution containing 0.02 M stochiometric Ca-EDTA complex (we assume that the pH is ideal, T = 25 °C). KCa-EDTA = $5x10^10$.

A) $5.0x10^-3$ M

B) $1.0x10^-2$ M

C) $6.3x10^-7$ M

D) $2.0x10^-2$ M
\end{apxtcolorbox}

%% file: appendix/prompt_optimization.tex
\subsection{Tasks}
\label{appendix:prompt-optimization-tasks}
Below, we provide an example query for each of the tasks in the prompt optimization section. 

\begin{apxtcolorbox}[Example Query for Word Sorting]
    Sort the following words alphabetically: List: oakland seaborg jacobi membrane trapezoidal allis marmot toggle anthology.
\end{apxtcolorbox}

\begin{apxtcolorbox}[Example Query for Object Counting]
    I have a couch, a bed, a car, a fridge, two tables, an oven, a toaster, and a chair. How many objects do I have?
\end{apxtcolorbox}

\begin{apxtcolorbox}[Example Query for GSM8k]
    Amber, Micah, and Ahito ran 52 miles in total. Amber ran 8 miles. Micah ran 3.5 times what Amber ran. How many miles did Ahito run?
\end{apxtcolorbox}

For word sorting and object counting, we obtain the datasets from the \href{https://github.com/suzgunmirac/BIG-Bench-Hard}{BBH repository}, and we randomly split examples into 50~(training)/100~(validation)/100~(test) samples. For GSM8k, we use the splits provided in DSPy~\citep{khattab2024dspy} which has 200~(training)/300~(validation)/1319~(test) samples.

\textbf{Evaluation:} For object counting and GSM8k, we use the string-based exact match metric, which looks at the last numerical value provided in the answer, and compares it to the ground truth answer. For word sorting, we prompt \gpto~to compare the ground truth list to the response provided in the answer, through the following prompt:

\begin{apxtcolorbox}[Evaluation system prompt for Word Sorting evaluation]
\textbf{System Prompt:}
Below is a question from a question-answering task, the ground truth answer, and reasoning with the final prediction. Is the final prediction correct, i.e. the same as the ground truth answer? Say only 1 (yes) or 0 (no). Return your response within <ACCURACY> </ACCURACY> tags. e.g.<ACCURACY> 0 </ACCURACY> or <ACCURACY> 1 </ACCURACY>.

\vspace{1em}

\textbf{Example prompt:}

**Question for the task:** \{question\}

**Ground truth answer:** \{answer\}

**Reasoning and prediction from the language model:** \{prediction\}
\end{apxtcolorbox}

%% file: appendix/molecule_optimization.tex
\subsection{Docking and Druglikeness Evaluation}
\label{molopt_evaluation}
To optimize molecules, we evaluate the binding affinity and druglikeness of chemical structures encoded as SMILES strings. To compute both metrics, the generated SMILES string is first converted into an octet-complete Lewis dot structure using RDKit's \texttt{MolFromSmiles} functionality. This method ``sanitizes'' molecules by adding explicit hydrogens, kekulizing aromatic rings, standardizing valence states, and assigning radicals \cite{rdkit}. If this sanitization process fails at any step, whether through a structural ambiguity or an invalid molecule, the QED and Vina scores are replaced with a single string informing \textgrad~that the molecule is invalid. 

If the generated SMILES string does represent a valid chemical structure, we compute the QED score using RDKit's \texttt{Chem.QED} function. While QED scores can be quickly and reliably computed, docking scores can vary significantly depending on target and ligand structure preparation. To ensure consistency, we calculate Vina scores using the DOCKSTRING package \cite{garcia2022dockstring}, which implements a standardized docking workflow and provides a set of gold-standard targets. In particular, after the ligand has been sanitized by RDKit, DOCKSTRING (de)-protonates it at pH 7.4 using Open Babel, and prepares and refines the 3D geometric structure using the Euclidean distance geometry algorithm ETKG and the  classical force field MMFF94. Finally, Gasteiger charges are computed for all ligand atoms, and the resulting structure is saved as a ligand PDBQT file passed to Autodock Vina. For target preparation, the DOCKSTRING benchmark suite provides 58 curated crystal structures of clinically  relevant proteins, with the majority at less than 2.5 \AA~resolution. These structures are specially prepared to improve correlations between theoretical and experimental binding affinities, for example by manual addition of polar hydrogens and removal of residual water and solute molecules. DOCKSTRING also standardizes simulation parameters such as numerical seeds, search box coordinates, and sampling exhausitivity, to ensure reproducible scoring.  

\subsection{Objective Functions}
\label{molopt_objective_function}
Once the scores have been computed, they are passed to an LLM in the following format. 

\begin{apxtcolorbox}[Molecule Optimization Prompt]
Given a docking and a druglikeness score, and a molecule as a SMILES string provide a short criticism to improve the druglikeness of this molecule and its binding affinity with the protein \{\texttt{protein\_name}\}. For docking, lower is better (less than $-10$ is considered good) and for druglikeness, $1$ is the best and $0$ is the worst (greater than $0.8$ is considered good). In terms of prioritization, the docking score is \{\texttt{vina\_qed\_ratio}\} times as important as the druglikeness score. Make sure your criticism is very succinct and to the point. \\

\# \texttt{if smiles\_string is valid} \\
SMILES: \{\texttt{smiles\_string}\}, Docking: \{\texttt{Vina}\}, Druglikeness: \{\texttt{QED}\} \\

\# \texttt{if smiles\_string is invalid} \\
SMILES: \{\texttt{smiles\_string}\}, This molecule is invalid. 
\end{apxtcolorbox}

Note that this prompt allows us to specify both the \textit{target name} as well as a \textit{prioritization} between these two objectives. When optimizing promising molecules in late stage drug discovery, medicinal chemists use detailed structural knowledge about a protein target's binding pocket in addition to docking scores, for example through interactive $3$D molecular visualization software. To preserve the generality of \textgrad~in our experiments, we do not to attempt to provide similar geometric information, as not all LLMs support multimodal inputs. However, we do include the \texttt{protein\_name} in the loss prompt to inject supplementary structural information. 

In practice, drug efficacy is typically a higher priority than absorbtion efficiency \cite{hughes2011principles, wenlock2003comparison}, so we set the \texttt{vina\_qed\_ratio} to be $10$. Empirically, we observe that by scaling this prioritization factor, we can tune \textgrad's generation towards molecules with differing binding affinity and druglikness tradeoffs. This prioritization also allows us to simplify post-hoc selection and ranking of the generated molecules with a single ``overall'' score defined as follows, where a lower overall score $s_{\texttt{overall}}$ indicates a better molecule. 
\begin{align}
    s_{\texttt{overall}}(\text{molecule}, \text{protein}) = \texttt{Vina}(\text{molecule}, \text{protein}) + (1-\texttt{QED}(\text{molecule}))
    \label{eq: molopt_summary_score}
\end{align}
Since the QED score is bounded between $0$ and $1$, and the Vina score typically ranges between $-3.0$ to $-12.0$ kcal/mol, this overall score places approximately $10$ times more emphasis on binding affinity than druglikeness. 

\subsection{Benchmarks}
\label{molopt_benchmark}
To benchmark the performance of \textgrad~generated molecules, we compare their characteristics to clinically approved drugs for the same protein targets found in DrugBank, a database of $16,619$ drugs \cite{knox2024drugbank}. To ensure that the DrugBank molecules were both comparable and high quality, we filtered for drugs that were small molecules, had full clinical approval, and were designed for orthosteric binding with the same active site in the DOCKSTRING benchmark suite. After these filtering criteria, we identified $118$ drugs targeting $29$ of the $58$ DOCKSTRING proteins. When evaluating binding affinity and druglikeness for the clinically approved drugs, we compute the Vina and QED scores using the same tools and workflow described in section \ref{molopt_evaluation}, exactly as applied to the \textgrad~molecules.

\subsection{Initialization}
\label{molopt_initialization}
\begin{figure}[!htb]
    \centering
    \includegraphics[width = 0.8\linewidth]{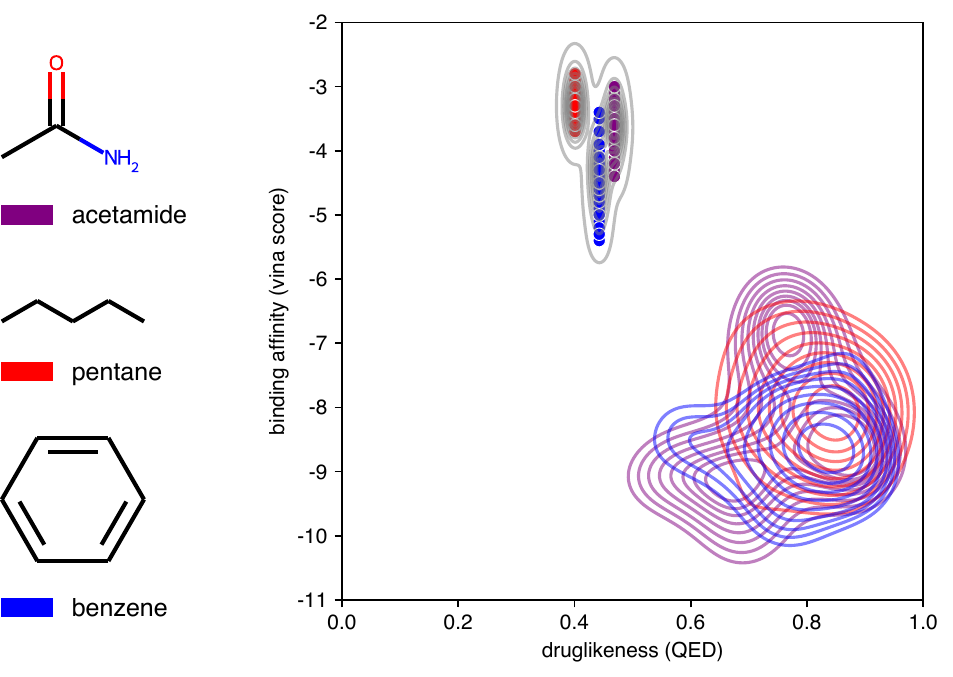}
    \caption{\textbf{Molecule Initialization}. We initialize \textgrad~with fragment molecules from three, diverse functional groups and optimize each initial molecule for $10$ iterations for all $58$ targets in the DOCKSTRING benchmark suite. For each fragment and each protein, we perform post selection on the generated molecules using the summary score outlined in equation \ref{eq: molopt_summary_score} and visualize the resulting distribution. We observe that while the initial QED and Vina score distributions of the starting fragments varies greatly, the distribution of the optimized molecules is highly overlapping.}
    \label{fig:molecule_initialization}
\end{figure}

In practice, molecular optimization is typically accelerated by large scale pre-optimization screening, where libraries of millions or even billions of existing chemical structures are scored and ranked by docking, druglikeness, and other metrics. Only the most promising structures, termed "leads", are further refined by medicinal chemists \cite{berry2015practical}. While \textgrad~is capable of optimizing any initial molecule, in this work, to more accurately characterize it's performance and avoid biasing its designs towards existing drugs, we instead select our three initial molecules from simple fragments of common functional groups, shown in Figure \ref{fig:molecule_initialization}. These fragments are highly diverse, and belong to different functional groups. Although these initial fragments have differing druglikness and binding affinity characteristics, the impact of the starting fragment on \textgrad's performance is minimal, and \textgrad~is capable of generating molecules with high druglikeness and binding affinity from all three fragments. 

\subsection{Chemical Novelty}
\label{molopt_novelty}
\begin{figure}[!htb]
    \centering
    \includegraphics[width = \linewidth]{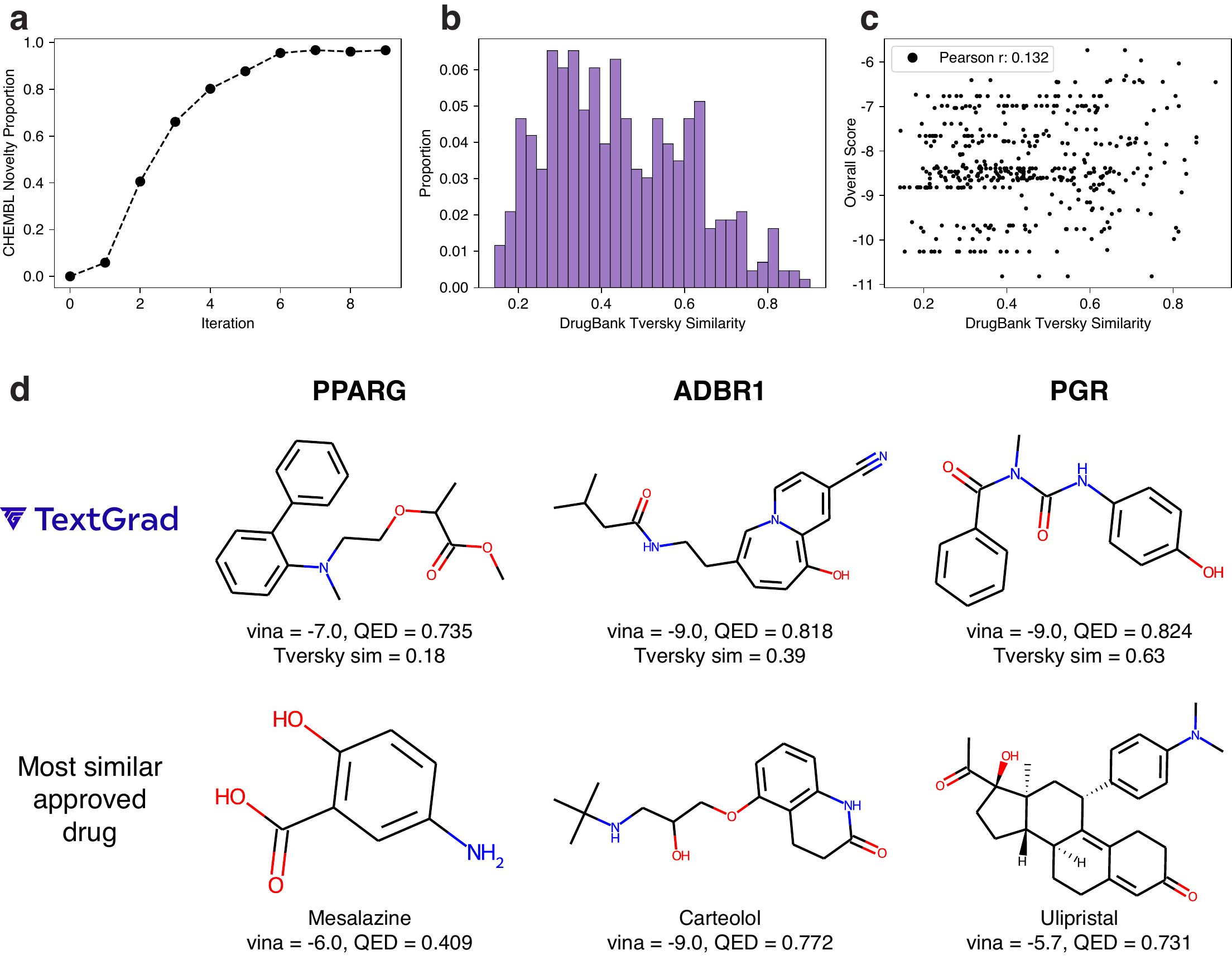}
    \caption{\textbf{Structural Novelty} In panel \textbf{(a)}, we observe increasing novelty over optimization updates, where a generated molecule is considered``novel'' if there does not exist any existing molecule in ChEMBL with a  Tanimoto similarity score greater than $0.8$. In panels \textbf{(b,c,d)}, we estimate substructure similarity using the Tversky metric between \textgrad~molecules and clinically approved drugs for each target. While there are a wide range of similarity scores, we observe that similarity with known drugs has little correlation with molecule performance, suggesting that \textgrad's generation process is weakly influenced by ``memorized'' knowledge of clinically approved structures.}
    \label{fig:molecule_novelty}
\end{figure}

One of the primary concerns with using LLM models in scientific discovery is their capacity to simply memorize and regurgitate their training sets instead of performing logical reasoning \cite{birhane2023science}. Since recent models like \gpto~are trained on massive, opaque datasets, a key concern is that the molecules \textgrad~generates may simply be duplicates of clinically approved molecules for their respective targets. To quantify and test the extent or existence of this ``memorization'' effect, we compare the molecules that \textgrad~generates with approved drugs for their respective protein targets, as well as unrelated but existing druglike chemical compounds. 

We first compare \textgrad~generated molecules for a particular protein target with all the clinically approved small molecules found in Drugbank for the same target (see \ref{molopt_benchmark} for filtering criteria) using the Tversky similarity score on RDKit daylight chemical fingerprints. The Tversky similarity score is not symmetric, and compares substructures between a variant and reference molecule \cite{nikolova2003approaches}. In our application, this is preferred as it assigns a high score to generated (variant) molecules that contain most or all of the substructures in the Drugbank (reference) molecule, even if there exist other extraneous substructures in the generated molecule that reduce the symmetric overlap between the two molecules. A Tversky score of $1.0$ indicates that the \textgrad~molecule is a complete subset of the Drugbank molecule, while $0.0$ indicates no overlap. 

In order to analyze the most relevant chemical structures, we restricted our analysis to a set of high performing \textgrad~molecules, rather than all molecules from all iterations. In particular, using the overall score described in Equation \ref{eq: molopt_summary_score}, we selected the best molecule for each protein target and for each initial fragment, for a total of $87$ generated molecules ($29$ druggable targets x $3$ initial fragments) and $118$ clinically approved drugs across $29$ targets. For each target, we compute all pairwise Tversky scores between the $3$ \textgrad~molecules and the Drugbank molecules approved for that target, setting the Drugbank molecule as the reference, and the \textgrad~molecule as the variant. 

We observed that the distribution of Tversky similarity scores was quite broad, with a median of $0.42$ but ranging from  $0.14$ to $0.90$ (Figure \ref{fig:molecule_novelty} \textbf{(b)}). However, we observe that Tversky similarity between \textgrad~and DrugBnak molecules is actually slightly anti-correlated with molecule performance as measured by the overall score, suggesting that \textgrad~'s optimization procedure is at best weakly influenced by prior knowledge of approved drugs. In fact, for a variety of generated molecules along a gradient of similarity scores, the generated molecules exhibit QED and Vina scores that match or even exceed their DrugBank counterparts (Figure \ref{fig:molecule_novelty} \textbf{(d)}). 

Beyond the similarity to known drugs, we are also interested in observing if \textgrad~ is generating novel molecules, or discovering previously unknown properties in existing compounds, for example a strong binding affinity to a protein target in a compound that has not been associated with the protein. To answer this question, we perform a Tanimoto similarity search across all of ChEMBL, a manually curated database of $2.4$ million bioactive molecules with drug-like properties \cite{gaulton2012chembl}. Unlike the Tversky score, the Tanimoto metric measures the symetric overlap between two chemical structures, where $1.0$ indicates an exact match, and $0.0$ no overlap. We classify a \textgrad~compound as ``novel'' if and only if there does not exist any molecule in ChEMBL with a with a Tanimoto similarity score over 0.80. 

While the starting fragments are known molecules, we observe that as the number of iterations increases, \textgrad~generates molecules that are progressively less likely to be previously known compounds. By the $6$th iteration, 95\% of all the molecules generated by \textgrad~across all $58$ targets and $3$ starting fragments are novel using the criteria above (Figure \ref{fig:molecule_novelty} \textbf{(a)}). By observing the trajectories of generated molecules and analyzing the textual gradients, we hypothesize that the feedback provided by the Vina and QED scores encourages \textgrad~to explore chemical space beyond known molecules by progressively adding and removing combinations of functional groups. Since these chemical updates can form a combinatorial number of unique structures, it is reasonable that \textgrad~would reach previously unexplored regions of chemical space in a relatively small number of iterations.   

\subsection{Implicit Objectives}
\label{molopt_admet}

\begin{figure}[!htb]
    \centering
    \includegraphics[width = \linewidth]{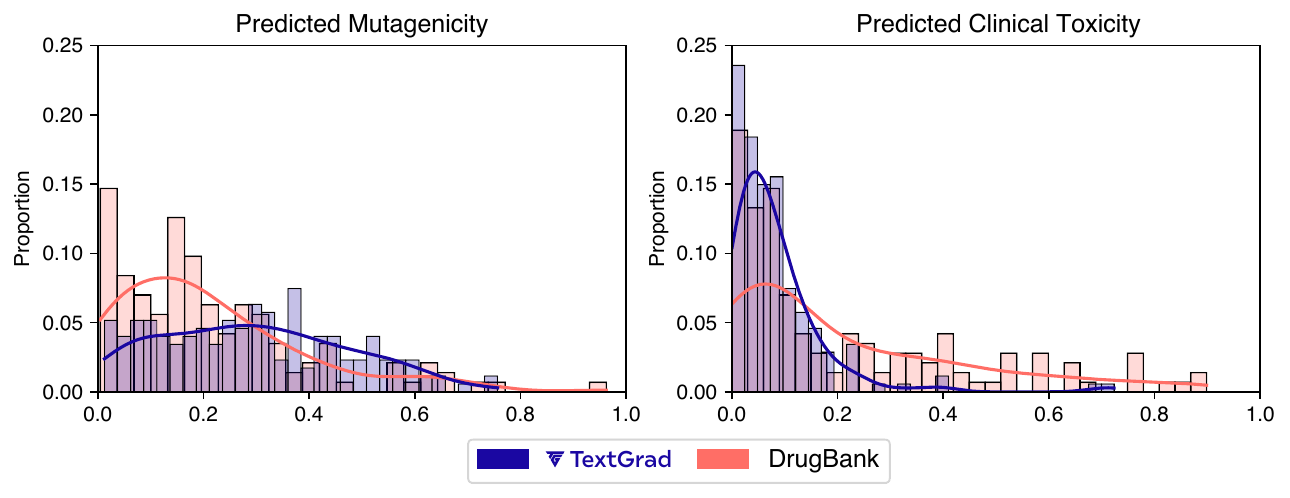}
    \caption{\textbf{Safety Properties} We evaluate the predicted harmfullness of \textgrad~molecules using the ADMET-AI model and compare them to clinically approved drugs. For both  mutagenicity and clinical toxicity, $1.0$ indicates a highly likelihood for harm and $0.0$ a low likelihood. We observe that despite the fact that neither of these characteristics are directly encoded into \textgrad's objective function, \textgrad~implicitly avoids proposing harmful molecules.}
    \label{fig:molecule_admet}
\end{figure}

Another key concern in applying LLMs in science is their propensity for hallucinations, where models generate factually  incorrect or illogical responses while attempting to satisfy user requests \cite{birhane2023science}. In our setting, this hallucinations could manifest by \textgrad~proposing invalid, toxic, or otherwise undesirable molecules in order to optimize its objective function. We control for severe hallucinations by preprocessing molecules using RDKit sanitization, ensuring at the bare minimum that \textgrad~generates chemically valid molecules. However, this simple preprocessing step does not completely specify desirable chemical behavior. A direct strategy would be to exhaustively encode all possible metrics for desirability in drug molecules beyond druglikeness and binding affinity into the objective function, and extend it to include sythesizability, toxicity, among other criteria. Unfortunately, this approach is not realistically feasible as not all criteria for desirability have mature computational metrics. Thus, a key question is whether \textgrad~obeys so-called \textit{implicit} objectives during its optimization process that curtails illogical or undesirable behavior. 

To evaluate the extent or existence of undesirable molecules, we can characterize the harmfulness of the generated molecules, focusing on mutagenisis and clinical toxicity. Mutagenicity refers to the ability of a drug to induce genetic alterations, which may lead to DNA damage and harmful long-term affects. Clinical toxicity refers to a broad range of adverse short or long-term side effects. Importantly, neither druglikeness nor binding affinity are strongly correlated with these criteria, and thus these desirability metrics are not directly optimized or otherwise explicitly encoded in \textgrad's objectives. 

To evaluate the propensity for mutagenisis and clinical toxicity, we employ the ADMET-AI model, that predicts these scores from the chemical structures of molecules \cite{swanson2023admet}. ADMET-AI employs a deep learning model trained on multiple relevant datasets. In particular, for mutagenicity, ADMET-AI is trained on $7,255$ drugs from the Ames dataset, a bacterial reverse mutation assay for rapidly screening large numbers of compounds for can induced genetic damage and frameshift mutations. A predicted label of $1.0$ indicates a high likelihood that the drug will induce mutagenesis, while a label of $0.0$ indicates a low likelihood. For clinical toxicity, ADMET-AI is trained on the ClinTox dataset, a dataset of $1,484$ drugs consisting of molecules that have failed clinical trials for toxicity reasons and also drugs that are associated with successful trials. Similarly, a predicted label of $1.0$ indicates a high likelihood of clinical toxicity, while a label of $0.0$ indicates a low likelihood. 

Once again, we restrict our analysis to the best performing generated molecules with druggable targets as measured by the overall score, and select the best molecule for each protein target and for each initial fragment, for a total of $87$ generated molecules. We then compare their predicted mutagenisis and clinical toxicity to the the $118$ clinically approved molecules from DrugBank. We observe that for both Mutagenicity and Clinical Toxicity, the molecules generated by \textgrad~have predicted distributions that indicate a low likelihood of harmful effects, and closely match the distributions of the clinically approved molecules. Together, these results suggest that \textgrad~implicitly avoids proposing harmful molecules, even though these criteria are not directly encoded in its loss function.

%% file: appendix/treatment_plan.tex
\subsection{Prompts}
Radiotherapy treatment plan evaluation can based on various dimensions, therefore there is no single score that can indicate the quality of plans. We adopt LLM to compute the ``loss'' by prompting it to assess the plan quality with clinical protocols. Specifically, LLM is used to compare each protocol with the current plan and produce the final assessment.
\label{tp_prompts}
\begin{figure}[!t]
\begin{apxtcolorbox}[Treatment Plan Loss Prompt]
Please act as an impartial and objective professional radiation oncologist. Your job is to evaluate the quality of radiation therapy treatment plans based on their dose volume histogram. You should consider the following protocols: 

<Clinical Protocols>
    
Note that the order above does not indicate priority; always prioritize the regions that have protocols that are more significantly violated.

Here is the dose-volume histograms of the candidate plans for evaluation; each entry in the dose-volume histograms (DVH) table indicates the percentage of volume receiving a dose higher than a certain Gy (specified in the first column). 

<DVH table>

We also provide the statistics of the above DVH table for ease of evaluation. 

<DVH statistics>

Now, based on the protocols, and the DVH, please evaluate the plans. Avoid any positional biases and ensure that the order in which the responses are presented does not influence your decision. Be as objective as possible. Your answer must: 
\begin{enumerate}
    \item (Evidence) Extract the corresponding entries from the DVH table based on the factors to consider.
    \item (Interpretation) Based on the extracted entries from DVH table, interpret them as: \texttt{\{percentage\}\%} of volume encompassing the dose more than \texttt{\{dose\}} Gy for \texttt{\{organ\}}.
    \item (Analysis) Combine the interpretation and the factors, analyze whether each of the numerical factors are met.
\end{enumerate}
Based on the analysis on DVH tables, you need to produce a final evaluation. When all factors are satisfied, you should always answer no improvement is needed. If improvements are required, please provide suggestions on where to improve. Please ensure to follow the format below to return the evaluation results:\\
\\
<FINAL> Decision: [The plan does/doesn't need to be improved.] Reasons: [list all factors that are not satisfied with detailed reasons, e.g. protocol X on (PTV/rectum/bladder/fh/body) is not satisfied because ...] </FINAL>\\
\\
The final answer in the end must strictly follow the format above.

\end{apxtcolorbox}
\end{figure}
\subsection{Inner-loop optimization for treatment planning}

We employ a two-loop optimization approach \cite{xingOptimizationImportanceFactors1999}, which includes (i) an inner loop for inverse planning and (ii) an outer loop for optimizing the hyperparameters of the inner loop. The inner loop focuses on traditional fluence map optimization, seeking to determine the optimal fluence map $x$ by minimizing a cost function that combines multiple weighted objectives for various targets and organs at risk. This cost function is defined as:

\begin{equation}
    \begin{aligned}
        \min_{x} \quad & \sum_{t=1}^{N_t} w_{t}([Kx]_t - d_{t})^2 + \sum_{s=1}^{N_s} w_{s}\Theta([Kx]_s - d_{s})([Kx]_s - d_{s}) \\
        \textrm{s.t.} \quad &D_{95}([Kx]_t) = d_{t}, \\
          & x \geq 0, \\
    \end{aligned}
    \label{eq:inner}
\end{equation}

Here, \(\{w_t\}_{t=1}^{N_t}\) and \(\{w_s\}_{s=1}^{N_s}\) are the importance weights (the hyperparameters optimized by \textgrad) that balance the various objectives for \(N_t\) PTV targets and \(N_s\) OARs, respectively. \(K\) denotes the dose influence matrix, which specifies the dose per fluence unit delivered to each voxel in the volume by each beamlet. \(\{d_t\}_{t=1}^{N_t}\) and \(\{d_s\}_{s=1}^{N_s}\) are the scalar objective doses for each structure. The cost function essentially penalizes squared deviations from the target objective doses for the PTV targets and penalizes squared overdosing for the OARs only when doses exceed \(d_s\). The Heaviside function \(\Theta\) is used to ensure the objective considers only positive values. The minimization is constrained to positive \(x\) values and ensures that \(D_{95}\) --- the minimum dose received by 95\% of the structure volume --- matches the prescribed dose for the clinical goal. We use \texttt{matRad}~\cite{wieser2017development} with interior point algorithms to solve the inner-loop optimization.

\subsection{Additional Experimental Details}
\paragraph{Dataset} The dataset used in this study comprised imaging and treatment plans for 5 prostate cancer patients who underwent intensity-modulated radiation therapy (IMRT). Available data for each patient includes CT scans, delineated
anatomical structures, and clinically approved treatment plans obtained via Eclipse\textsuperscript{\tiny\textregistered}. %

\paragraph{Method} As we mentioned in~\ref{sec:treatment-plan}, \textgrad~is used to optimize the hyperparameters (e.g., importance weights for PTV and OARs) of the inner-loop numerical optimizer that generates the treatment plan. This optimization is done using a variation of vanilla~\textgrad, i.e. ``projected gradient descent with momentum updates''.In particular, three prostate cancer treatment plans optimized by clinicians, along with their corresponding hyperparameters, are provided. These examples guide the updates of the hyperparameters. This procedure can be viewed as an analogy to projection, as the updated hyperparameters are ``softly projected'' onto a feasible set defined by the three in-context examples. Moreover, the historical hyperparameters and the textual gradients from past iterations, as an analogy to momentum, are also included in the prompts for updating the hyperparameters. This additional context helps refine the optimization process. The optimization will be stopped if the loss suggests all protocols meet, other wise, it will be stopped if the maximum number of iterations (we set it to 10) is reached.  

\paragraph{Initialization} The hyperparameters i.e. the importance weights are all initialized at 100 for different organs. The dose objectives are set to 70.20 for PTV, 0.00 for bladder and rectum, and 30.00 for femoral heads and body, and fixed during optimization.

\subsection{Additional Results}
In Supplementary Table~\ref{tab:ptv} and ~\ref{tab:oar}, we show additional results on comparing \textgrad~optimized plan with clinicians optimized plans. 

\begin{table}[H]
\caption{\textbf{PTV dose metrics}. Several dose metrics of the PTV target are displayed for all the clinical and TextGrad optimized plans, including the mean and minimum doses, as well as the $D_{95}$. For all the metrics, we include the average deviations from the clinical goal across 5 plans and the standard deviation in brackets. Values in bold represent the best for each PTV target.}

    \begin{tabular}{@{}llllll@{}}
    \toprule
    \textbf{Target} & \textbf{Method}   & \textbf{Mean dose [Gy]} & \textbf{Min dose [Gy]}  & \textbf{Max dose [Gy]}
    & \textbf{$\textbf{D}_{95}$ [Gy]}   \\ \midrule
    \multirow{3}{*}{PTV} & Clinical Goal & 70.20 & $\approx70.20$ & $\approx70.20$ & 70.20 \\ 
                             & Radiation Oncologist & +1.97 (0.36) & -8.88 (2.31) & +4.66 (0.82 )& -0.10 (0.15) \\
                             & \textgrad & \textbf{+0.51} (0.09) & \textbf{-8.48} (2.38) & \textbf{+3.63} (0.87) &  \textbf{+0.00} (0.00)  \\
                             \bottomrule
    \end{tabular}
    \centering
    \label{tab:ptv}
\end{table}

\begin{table}[H]
\caption{\textbf{Organs at Risk (OARs) dose metrics}. We show mean dose capturing OAR sparing. Lower values demonstrate better OAR sparing which is desirable, as this number indicates organs at risk, which should not get more than dosage than what is listed in the clinical guidelines. For all the metrics, we include the average mean dose across 5 plans and the standard deviation in brackets.}

    \begin{tabular}{@{}llllc@{}}
    \toprule
    \textbf{Organ} & \textbf{Method} & \textbf{Mean dose [Gy] $\downarrow$} & \textbf{$\textbf{D}_5$$\downarrow$} & \textbf{$\textbf{D}_{50}$$\downarrow$}\\ \midrule

    \multirow{2}{*}{Rectum} & Radiation Oncologist & 23.88 (6.45) & 64.26 (10.00) & 20.04 (5.50)  \\
                             & \textgrad & \textbf{17.18} (4.2) & \textbf{58.82} (18.81) & \textbf{9.54} (0.70)\\ \midrule
    \multirow{2}{*}{Bladder} & Radiation Oncologist & 22.39 (5.55) & 67.81 (6.44) & 14.78 (8.42) \\
                            & \textgrad &  \textbf{20.92} (0.79) & \textbf{65.96} (6.96) & \textbf{14.11} (3.17) 
   \\ \bottomrule
    \end{tabular}
    \centering
    \label{tab:oar}
\end{table}